\documentclass[12pt,noinfoline,a4paper]{imsart}
\RequirePackage{amsthm,amsmath,amsfonts,amssymb}
\RequirePackage[authoryear]{natbib}
\RequirePackage[colorlinks,linkcolor=cyan,citecolor=red,urlcolor=magenta]{hyperref}
\usepackage[table,xcdraw]{xcolor}
\usepackage{arydshln}
\usepackage{color,soul}
\usepackage{accents,float,graphicx,subfig,multirow,dcolumn,booktabs,lscape}
\usepackage[a4paper,margin=2.54cm]{geometry}
\usepackage[ruled,linesnumbered]{algorithm2e}
\usepackage{comment}

\pagecolor[rgb]{0.1,0.1,0.1} %shallow black
\color[rgb]{0.85,0.85,0.85} %shallow grey

\startlocaldefs
\numberwithin{equation}{section}

\theoremstyle{definition}
\newtheorem{definition}{\protect\definitionname}
\providecommand{\definitionname}{Definition}
\numberwithin{definition}{section}

\newcommand{\ind}{\perp\!\!\!\!\perp}
\newcommand{\ts}{\rlap{$^{***}$}}
\newcommand{\ds}{\rlap{$^{**}$}}

\def\spacingset#1{\renewcommand{\baselinestretch}{#1}\small\normalsize}\spacingset{1}

\newcolumntype{.}{D{.}{.}{-1}}
\definecolor{red}{rgb}{1.0, 0.0, 0.0}

\endlocaldefs

\begin{document}

\begin{frontmatter}

  \title{Instrument variable detection with graph learning : an application to high dimensional GIS-census data for house pricing\thanksref{T1}}

  \thankstext{T1}{Xu would like to thank Google Australia and NICTA for hardware and programming assistance in package optimization and development. Xu also would like to thank Dr. Peter Exterkate, Uni Sydney and Prof. A. Colin Cameron, UC Davis for their valuable advice.Fisher would like to acknowledge the financial support of the Australian Research Council grant DP0663477.}

  \begin{aug}
    \author{\fnms{Ning }\snm{Xu,}\ead[label=e1]{n.xu@sydney.edu.au}}
    \author{\fnms{Timothy C.G. }\snm{Fisher}\ead[label=e2]{tim.fisher@sydney.edu.au}}
    \and
    \author{\fnms{Jian }\snm{Hong}\ead[label=e3]{jian.hong@sydney.edu.au}}

    \address{School of Economics, University of Sydney\\ NSW 2006 Australia\\
    \printead{e1,e2,e3}}
  \end{aug}

  \begin{abstract}
    Endogeneity bias and instrument variable validation have always been important topics in statistics and econometrics. In the era of big data, such issues typically combine with dimensionality issues and, hence, require even more attention. In this paper, we merge two well-known tools from machine learning and biostatistics---variable selection algorithms and probablistic graphs---to estimate house prices and the corresponding causal structure using 2010 data on Sydney. The estimation uses a 200-gigabyte ultrahigh dimensional database consisting of local school data, GIS information, census data, house characteristics and other socio-economic records. Using "big data", we show that it is possible to perform a data-driven instrument selection efficiently and purge out the invalid instruments. Our approach improves the sparsity of variable selection, stability and robustness in the presence of high dimensionality, complicated causal structures and the consequent multicollinearity, and recovers a sparse and intuitive causal structure. The approach also reveals an efficiency and effectiveness in endogeneity detection, instrument validation, weak instrument pruning and the selection of valid instruments. From the perspective of machine learning, the estimation results both align with and confirms the facts of Sydney house market, the classical economic theories and the previous findings of simultaneous equations modeling. Moreover, the estimation results are consistent with and supported by classical econometric tools such as two-stage least square regression and different instrument tests. All the code may be found at \url{https://github.com/isaac2math/solar_graph_learning}.
  \end{abstract}

  \begin{keyword}
    \kwd{instrument selection}
    \kwd{endogeneity detection}
    \kwd{subsample-ordered least-angle regression}
    \kwd{lasso regression}
    \kwd{elastic net regression}
    \kwd{variable selection}
    \kwd{random graph}
    \kwd{grouping effect}
  \end{keyword}

\end{frontmatter}

%%%%%%%%%%%%%%%%%%%%%%%%%%%%%%%

\spacingset{1.5}

%%%%%%%%%%%%%%%%%%%%%%%%%%%%%%%
%%%%%%%%  INTRODUCTION %%%%%%%%
%%%%%%%%%%%%%%%%%%%%%%%%%%%%%%%

\section{Introduction}

Endogeneity bias has long been a problem in causal analysis and has for decades been a focus of research by statisticians, econometricians and biostatisticians. With the ongoing increases in dimensionality, the topic requires more attention than ever. On the one hand, we seem to have more information, which raises the potential to observe and rectify endogeneity bias by finding a valid instrumental variable (referred to as instrument for short); on the other hand, the problem is complicated by the curse of high dimensionality and the consequent complication of dependence structures. Thus, it is important to investigate how best to utilise high-dimensional data for endogeneity detection and instrument selection while minimizing the curse of dimensionality. In this paper, we combine two theoretically well-founded machine learning and biostatistical tools---variable selection and random graph estimation---and demonstrate that the combination performs admirably in endogeneity detection and instrument selection in the presence of ultrahigh dimensional data.

In causal analysis, \citet[246]{pearl2009causality} shows that there are three definitions of a valid instrument: graphical criteria, error-based criteria and counterfactual criteria, where the graphical criteria implies the error-based criteria. Classical regression analysis relies mostly on the error-based criteria. In econometrics, the instrument $\mathbf{z}$ is typically defined by the data-generating process
\begin{equation}
  \begin{cases}
    \mathbf{x} & = \alpha_0 + \alpha_1 \mathbf{z} + \mathbf{v}\\
    Y          & = \beta_0  + \beta_1  \mathbf{x} + \mathbf{u}
  \end{cases}
  \label{eqn:instrument}
\end{equation}
\noindent
where $\left\{ u, v \right\}$ are noise terms, $\left\{ \mathbf{z}, \mathbf{v} \right\}$ cause $\mathbf{x}$, $\left\{ \mathbf{x}, \mathbf{u} \right\}$ cause $\mathbf{y}$ and $\mathbf{x}$ is endogenous.\footnote{Unfortunately the causation assumption cannot be dropped; otherwise, endogeneity will inevitably arise. See Appendix~\ref{App:IV_def}.} For the validity of an instrument $\mathbf{z}$, we typically require

\begin{itemize}
  \item[\textbf{C1}] $\mathrm{corr} \left( \mathbf{z}, u \right) = 0$, and
  \item[\textbf{C2}] $\mathrm{corr} \left( \mathbf{z}, \mathbf{x} \right) \neq 0$ in the population.
\end{itemize}

\textbf{C1} implies that changes in $\mathbf{z}$ cannot affect $u$, further implying that changes in $\mathbf{z}$ cannot affect $\mathbf{y}$ via $u$. \textbf{C2} implies that changes in $\mathbf{z}$ can affect $\mathbf{x}$, further implying that changes in $\mathbf{z}$ can affect $\mathbf{y}$ via $\mathbf{x}$. \textbf{C1} and \textbf{C2} together mean that $\mathbf{z}$ can only impact $\mathbf{y}$ via $\mathbf{x}$.

\begin{figure}[H]
  \centering
  \includegraphics[width=0.2\paperwidth]{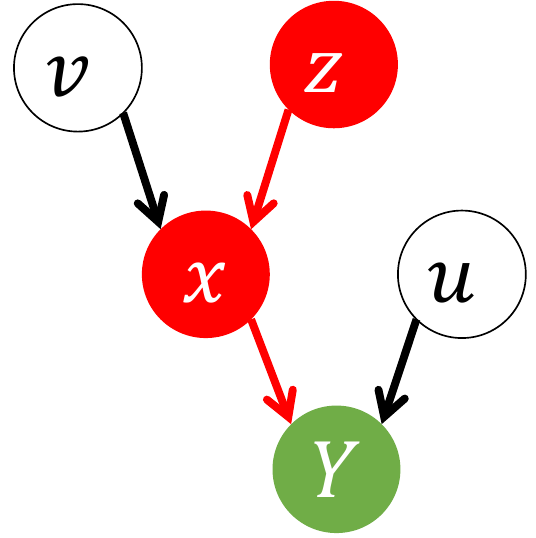}
  \caption{The graph representing~(\ref{eqn:instrument}). The absence of arrows between $\mathbf{z}$ and $\mathbf{u}$ reflects that $\mathrm{corr} \left( \mathbf{z}, u \right) = 0$.}
  \label{fig:example_dag}
\end{figure}

The idea of an instrument can be generalized using probabilistic graph models (also called Bayes nets or causal networks). In probabilistic graph models, the causal structure in~(\ref{eqn:instrument}) can be expressed equivalently by a directed acyclic graph (\textbf{graph} for short) as in Figure~\ref{fig:example_dag}.\footnote{There are several notation systems for graphs. Throughout the paper we follow the notation in \citet{koller2009probabilistic}.} In much research on causal inference (e.g., \citet[p44]{spirtes2000causation}), causal structure is directly defined using graphs that visually represent the causal relationships between variables.
\noindent
A useful analog to a graph is a family tree, where family members (variables) are connected by arrows representing parentage (causation). In Figure~\ref{fig:example_dag}, the arrows from $\{\mathbf{z}, \mathbf{v}\}$ to $\mathbf{x}$ mean that $\mathbf{z}$ and $\mathbf{v}$ \emph{directly cause} $\mathbf{y}$, further implying that $\mathrm{corr} \left( \mathbf{z}, \mathbf{x} \right) \neq 0$. Analogously, we say that $\{\mathbf{z}, \mathbf{v}\}$ are the \textbf{parents} of $\mathbf{x}$, $\mathbf{x}$ is the \textbf{child} of $\{\mathbf{z}, \mathbf{v}\}$ and $\mathbf{z}$ is a \textbf{spouse} of $\mathbf{v}$. In Figure~\ref{fig:example_dag}, $\mathbf{v}$ directly causes $\mathbf{x}$ and, hence, \emph{indirectly causes} $\mathbf{y}$. The variables that directly or indirectly cause $\mathbf{y}$ are the \textbf{ancestors} of $\mathbf{y}$. Hence, $\mathbf{y}$ and $\mathbf{x}$ are the \textbf{descendants} of $\mathbf{z}$. Lastly, two variables are \textbf{siblings} if they share the same parents.\footnote{See \citet[Section 2.2]{koller2009probabilistic} for further detail on the terminology.}

%#DONE:  stress that Examples 1-3 are there to explain the usefulness of graph learning as well as to illustrate common issues that arise in causality inference

In statistics and biostatistics, causal inference is typically conducted via two stages. The first stage is to estimate the causal structure of the key variable $\mathbf{y}$ (such as finding all its parents, children and spouses). Based on the estimated causal structure, we estimate the magnitude of the causal effects from $\mathbf{y}$ or to $\mathbf{y}$. As one of the most popular tool for structure estimation, graph learning provides unparalleled clarity on the causal structure specification. As a result, graph learning is typically considered critical for the accuracy and stability of the causal inference result. Failure to learn an accurate graph may result in different kind of estimation biases and cause a number of consequential problems, such as endogeneity, multicollinearity and misinterpretation. We illustrate this point using the following three examples.

\subsection*{Motivating examples}

Field knowledge and experience are frequently relied on when constructing the causal structure in empirical researches. However, if field knowledge is incomplete, unclear or partially misspecified, it may cause a series of problems for causal inference. In the following examples, we demonstrate that even a very small problem on causal structure specification can cause severe problems. Specifically, example 1 shows that correct causal inference relies on the temporal order of variable, e.g., when the value of a variable is determined.
\medskip

\begin{figure}[H]
  \centering
  \includegraphics[width=0.2\paperwidth]{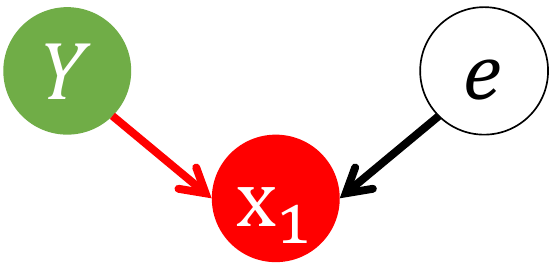}
  \caption{Example~1 dependence structure.}
  \label{fig:example_1}
\end{figure}

\noindent
\textbf{Example 1.} In both numerical and theoretical analysis of linear regression, typically $\mathbf{x}$ --- the parent of $\mathbf{y}$ --- is on the right-hand side of the equation and $\mathbf{y}$ is on the left. Such variable allocation is consistent with the order of time stamps --- the ancestor(s) on the right and its descendant on the left; moreover, it reflects the fact that, during the data generating procedure, $\mathbf{x}$ has to be generated first to determine the value of the child variable $\mathbf{y}$. However, in empirical analysis, field knowledge may fail to provide the detailed temporal order of variables. For example, suppose the true causal structure appears as in Figure~\ref{fig:example_1},

\noindent
reflecting the data-generating process
\begin{equation}
    \mathbf{x}_1 = \beta_0 + \beta_1 \mathbf{y} + e,
  \label{eqn:example_1x}
\end{equation}
where $\mathbf{x}_1$, $\mathbf{y}$ and $e$ are $n \times 1$ vectors; both $\mathbf{y}$ and $e$ cause $\mathbf{x}_1$; $\mathbf{y}$ is independent from $e$. Clearly, (\ref{eqn:example_1x}) implies that, in order to determin the value of $\mathbf{x}$, the values of $\mathbf{x}_1$'s parents --- $\mathbf{y}$ and $e$ --- must be determined first. Suppose that (i) the field knowledge confirms that $\mathbf{y}$ and $\mathbf{x}_1$ does have a causal relation; (ii) the field knowledge is not clear on which variable is generated first. Due to the fact that $\mathbf{y}$ is the variable of interests, $\mathbf{y}$ will typically be chosen as the response variable. Hence, the empirical model is
\begin{equation}
    \mathbf{y} = \alpha_0 +  \alpha_1 \mathbf{x}_1 + u
  \label{eqn:example_1Y}
\end{equation}
with $\alpha_0 = - \beta_0/\beta_1$, $\alpha_1 = 1/\beta_1$ and $u = - e/\beta_1$. Thus, $\mathrm{corr} \left( u, \mathbf{x}_1 \right) = \mathrm{corr} \left( - e/\beta_1, \mathbf{x}_1 \right) \neq 0$ and (\ref{eqn:example_1Y}) suffers from endogeneity. Worse, by mistaking the parent ($\mathbf{y}$) to be a child, the causal structure in (\ref{eqn:example_1Y}) is completely wrong. In this case, it would be very difficult to find an instrument to solve the problem. As a result, either the model is set up correctly by the temporal order, or it will be contaminated by endogeneity; there is hardly any middle ground. $\qed$
\medskip

Example 1 illustrates the importance of specifying the correct parent-child relation in a graph. Such problem referred to as \textbf{Markov equivalence} in graph learning, which will be detailedly discussed in latter sections. Statistical tools (like information criteria and tests) can verify whether there is likely to be a non-zero population correction between $\mathbf{x}$ and $\mathbf{y}$; however, without further information, they cannot identify whether $\mathbf{x}$ causes $\mathbf{y}$ or the other way around. To avoid such misspecification, one method is to collect the `time stamp' of each variable (when the value of a variable is determined) and order variables temporally. This requires that we should collect all the details about each variable. However, the time stamp issue is only part of the possible problems for inaccurate causal structure estimation. In Example 2, we demonstrate that, without an accurate graph, the causal inference procedure may be misled into a wrong track and never returns the true result.
\medskip

\noindent
\textbf{Example 2.} In regression analysis, it is well-known that regression coefficient estimates will be biased and inconsistent if an important covariate is omitted. To avoid omitted variable bias, it is often recommended (for example, \citet{pratt1988interpretation}) that investigators enlarge the set of potential covariates and control more variables. In this example, we demonstrate that this may also cause problems in causal inference.

Assume that the data are generated by the causal structure in Figure~\ref{fig:example_dag} and that the data generating process is (\ref{eqn:instrument}). Suppose that we want to investigate the causal effect from $\mathbf{z}$ to $\mathbf{y}$. Unfortunately, if we include $\mathbf{x}$ in the regression equation to avoid omission variable bias and set the equation as
\begin{equation}
  \mathbf{y} = b_0 +  b_1 \mathbf{x} +  b_2 \mathbf{z} + e,
  \label{eqn:example_2}
\end{equation}
we may never have an accurate inference on the causal effect from $\mathbf{z}$ to $\mathbf{y}$. As shown in Figure~\ref{fig:example_dag}, $\mathbf{z} \rightarrow \mathbf{x} \rightarrow Y$, implying there is an indirect causal relation from $\mathbf{z}$ to $\mathbf{y}$ via $\mathbf{x}$. However, if we include $\mathbf{x}$ in our regression, the value of $\mathbf{x}$ will be controlled when you investigate the relation between $\mathbf{z}$ to $\mathbf{y}$. This implies that $\mathbf{z}$ cannot affect the value of $\mathbf{x}$ and, hence, $\mathbf{y}$. As a result, $\mathbf{z}$ will not be significant in the regression equation and we may wrongly conclude that $\mathbf{z}$ has no causal effect on $\mathbf{y}$.

Worse, in empirical analysis, the culprit is likely to be mistaken from the persepctive of multicollinearity rather than a wrong causal structure. Noticing the high correlation between $\mathbf{x}$ and $\mathbf{z}$, some may wonder whether there is an omitted confounder for $\mathbf{x}$ and $\mathbf{z}$, resulting in an even larger control variable set; some may be misled and focus on improving the robustness of the regression instead of reconsidering the causal structure that the regression equation implies. Essentially, an increase in sample size or regression with robust standard errors may address multicollinearity issue only if the underlying causal structure of the regression equation is correct. By contrast, the source of the problem here is that `we control a variable that we should not'. In this case, if we want to use single-equation OLS to correctly measure the magnitude of the causal effect from $\mathbf{z}$ to $\mathbf{y}$, the only solution is that we remove $\mathbf{x}$ from the equation.$\qed$
\medskip

Example 2 clearly reveals that, similar to variable omission, redundant variables may also be a severe problem for the causal effect estimation. Put it another way, "control a wrong variable" and "control too few" are both problematic. To avoid the latter and the consequential omission variable bias, we typically focus on enlarging the set of control variables; however, such strategy is meaningful only if we know the correct causal structure in advance, which will prevent controlling a `wrong' variable like example 2. As a result, a systematic causal structure estimation needs to be done before measuring the magnitude of the causal effect in OLS, implying that a careful selection of variables is necessary.

Moreover, different sets of control variables are required under different scenarios. In example 2, you only need to include the parents of $\mathbf{y}$ if you are interested in the direct (causal) effect to $\mathbf{y}$; by contrast, you need to drop all the parents of $\mathbf{y}$  from the regression equation if you aim to measure the indirect effect from $\mathbf{y}$'s grandparents. This implies that, to well serve causal inference, an accurate graph is highly recommended. If learning the graph globally is impossible, at least we need variable selection algorithms to identify $\mathbf{y}$'s parents, spouses and children (whom $\mathbf{y}$ directly causes or is directly caused by). As a result, careful use of the selection algorithm is required well, especially for high dimensional data with complicated causal structure. Otherwise, variable selection algorithm may be misled and produce wrong causal structures, which may render the estimated structure unusable. Using the popular lasso algorithm in example 3, we demonstrate numerically the difficulty of correctly identifying the parents of $\mathbf{y}$ in the presence of a strong confounding effect and demonstrate how it causes problems on causal structure estimation. For precision and conciseness, we follow \citet{zhaoyu06, tibshirani2012strong} and quantify the difficulty of correctly identifying parents with the well-known irrepresentable condition (IRC).

\begin{figure}[H]
  \centering
  \includegraphics[width=0.35\paperwidth]{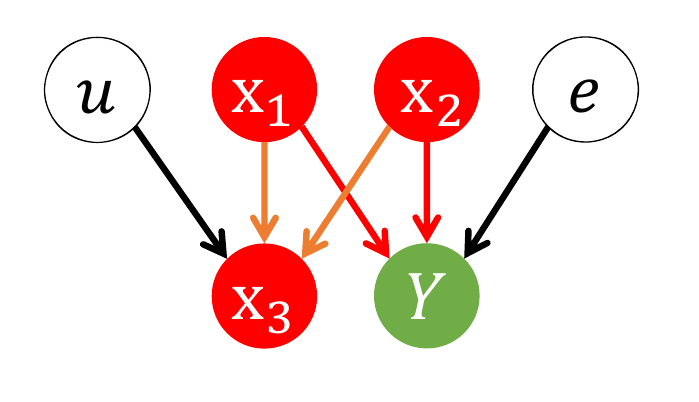}
  \caption{Example~3 causal structure.}
  \label{fig:example_2}
\end{figure}

\noindent
\textbf{Example 3.}\citep{zhaoyu06} Assume the data-generating process is
\begin{equation}
  \begin{cases}
    \mathbf{x}_3 = \omega_1 \mathbf{x}_1 + \omega_2 \mathbf{x}_2 + \sqrt{1 - \omega_1^2 - \omega_2^2} \; u, \\
    Y = \beta_1 \mathbf{x}_1 +  \beta_2 \mathbf{x}_2 + \sqrt{1 - \beta_1^2 - \beta_2^2} \; e, \\
  \end{cases}
  \label{eqn:example_3}
\end{equation}
where all variables are Gaussian; $\mathbf{x}_1$, $\mathbf{x}_2$, $\mathbf{x}_3$, $\mathbf{y}$, $e$ and $u$ are standardized $n \times 1$ vectors; $u$ and $e$ are independent from $\left\{u, e\right\}$. The causal structure shows that $\left\{\mathbf{x}_1, \mathbf{x}_2\right\}$ are the common parents of $\left\{\mathbf{x}_3, Y\right\}$, which are siblings. In this example, we try to use lasso to find the parents of $\mathbf{y}$.

The IRC states that, for variable selection accuracy in lasso (i.e., selecting $\left\{\mathbf{x}_1, \mathbf{x}_2\right\}$ and dropping $\mathbf{x}_3$ in this case), $\sum_i \left\vert \omega_i \right\vert < 1$. Otherwise, with a large probability, lasso-type estimators will take the sibling of $\mathbf{y}$ to be a parent (see the last simulation in \citet{ning2019solar} for detail). Worse, if a group of variables are highly correlated with one another, \citet{zou2005regularization} shows that lasso may randomly drop variables from the group (referred to as the \textbf{grouping effect}), making the variable selection process extremely sensitive to sampling randomness. As a result, lasso may wrongly include the sibling $\mathbf{x}_3$ as a parent or dump the true parent $\mathbf{x}_1$ and $\mathbf{x}_2$.

The consequnce of a wrong selection result is well beyond the variable redundancy in regression equations. There will be carried-on erros that can render the whole structure misspecified. In this example, if the variable selection algorithm mistake the sibling of $\mathbf{y}$ as its parent, all the children of the $\mathbf{y}$ --- the true nephews of $\mathbf{y}$ --- are consequently mistaken as $\mathbf{y}$'s siblings; if the variable selection algorithm mistake a parent of $\mathbf{y}$ (say $\mathbf{x}_1$) redundant, all the ancestors of the $\mathbf{x}_1$ --- the true ancestors of $\mathbf{y}$ and the variables indirectly causing $\mathbf{y}$ --- are consequently mistaken as redundant as well.

The ultimate consequnce is that, if a causal structure is misspecified, the succeeding statistical decisions regarding $\mathbf{y}$ will all be biased, such as selecting instrument variables, estimating the magnitude of a causal effect, prediction and forecast. Moreover, this also leads to difficulties with model interpretation and understanding. $\qed$
\medskip

As shown in example 3, we need to mitigate the caveats of traditional lasso estimator in empirical applications with severe multicollinearity or a complicated causal structure. To improve variable selection accuracy and robustness, we follow \citet{ning2019solar} and apply the novel subsample-ordered least-angle regression algorithm (\textbf{solar}) instead. Solar is derived from least-angle regression \citep{efronall04} and significantly outperforms lasso in terms of sparsity and variable-selection accuracy on data with severe multicollinearity and complicated causal structures. In particular, \citet{ning2019solar} shows that, unlike lasso, solar maintains variable selection robustness when IRC fails.

All previous examples shows that causal inference requires well thoughts, detailed `big data' with enough variables and time stamps, an accurate causal structure and a robust variables selection result that tolerates high dimensionality and multicollinearity caused by complicated causal structures. In this paper, we carefully pick all the tools that satisfy the requirements above and assemble them as a data-driven method to select and validate instrument variables  in high dimensional data.

\subsection{Literature review on graphical causal inference, graph learning and variable selection}

\subsubsection*{Causal inference based on probabilistic graphs learning}

Implementing a data-driven causal inference has been a central topic in machine learning and biostatistics for decades. \citet{verma1990causal} pioneered the use of graphs to analyze causal structure in the 1980s and summarize all the corresponding researches in \citet{pearl2009causality}. Building on that, \citet[p197]{spirtes2000causation} consider causal inference from the joint perspectives of graph learning and regression analysis. Both \citet{pearl2009causality} and \citet{spirtes2000causation} investigate the definition of instrument variables from the perspective of graph and show that it is implies that classical definition based on regression error. They illustrates that, as a special case of causal analysis, (i) the classical linear regression model typically assumes an oversimplified causal structure; (ii) regression can be easily misled by the complexity of causal structures in real-world data. As a result, it may not be reliable to estimate the magnitude of the causal effect using linear regression without verifying the causal structures. Overall, to correctly estimate the causal effect magnitude, these researches recommend graphs for learning causal structure data-driven in advance. Building on that, classical machine learning and biostatistics researches (for example, \citep{brito2002generalized, kuroki2005instrumental, chu2013semi, silva2017learning}) show that joint distributions alone are insufficient to determine whether an observable variable is a valid instrument, which is affected by a phenomenon called \textbf{Markov equivalence} and can be solved by finding the time stamps of each variable. Also, without specifying the time stamps, the researches still show that instrument variable assumptions can nevertheless be falsified by exploiting constraints in the joint distribution of multiple observable variables. The relevant algorithms are combinatorial and consequently requires huge computation loads even though the dimensionality of the data is not large. As a result, graphically causal inference would benefit deeply from a quick and accurate graph learning algorithm.

Graph learning are typically based on two methods : constraint-based learning and score-based learning (see, e.g., \citep{scutari2014bayesian}). To find a correct graph, constraint-based learning assumes a distribution on the joint distribution of all variables and carries out all conditional and marginal dependence tests among every possible pairs of variables. When testing the dependence between two variables, the constraint-based learning is senstive to the variables conditioned on. Also, this methods is combinatorially exhaustive and typically cost great computation loads when the dimensionality is large. By constrast, score-based learning computes an information criterion score (such as AIC, BIC, BGE score) for a given graph, selecting the graph with the minimal information criterion score. Score-based learning can be carried out using different packages (e.g., the \textsf{R} package \texttt{bnlearn} or the \textsf{Python} package \texttt{pgmpy}). Based on score-based and constraint-based learning, different algorithms are created. Typical constraint-based algorithms include Peter-Clark (PC) and Fast Causal Inference (FCI) \citep{spirtes2000constructing}. PC assumes that there is no confounder (unobserved direct common cause of two variables), and its discovered causal information is asymptotically correct. FCI gives asymptotically correct results even in the presence of confounders. Such approaches are widely applicable because they can handle various types of data distributions and causal relations, given reliable conditional independence testing methods. However, they do not necessarily provide complete causal information because they output Markov equivalence classes, i.e., a set of causal structures satisfying the same conditional independences. The PC and FCI algorithms produce graphical representations of these equivalence classes. In cases without confounders, there also exist score-based algorithms that aim to find the causal structure by optimizing a properly defined score function. Among them, Greedy Equivalence Search (GES) \citep{chickering2002optimal} is a well-known two-phase procedure that directly searches over the space of equivalence classes. To reduce the computation load of score-based learning, researchers in machine learning and biostatistics typically assume the jointly distribution is Gaussian and the dependence among variables are all linear, which implies a linear Gaussian graph (e.g., \citet{bollen1989structural, geiger1994learning, spirtes2000causation}) and later on is generalized as a linear non-Gaussian graph (e.g., \citet{shimizu2006linear}) and a nonlinear non-Gaussian graph (e.g., \citet{hoyer2008nonlinear}).

\subsubsection*{Variable selection and corresponding issues in linear graph learning}

In empirical researches, a graph learning task typically starts from linear graph learning (e.g., \citet{bollen1989structural, geiger1994learning, spirtes2000causation, friedman2008sparse}). Compared with nonlinear graphs on the same dataset, linear graphs requires much less computation load and is easier for inference. Also, linear graph is deeply related to linear modelling methods like linear regression, best subset variable selection, shrinkage and lasso. Only if linearity cannot approximate the dependency in the data, researchers apply nonlinear dependence measure like Hilbert Schmidt independence criterion \citep{gretton2005measuring} and mutual information.

As a major issue in linear graph learning, multicollinearity is frequently observed among variables with complicated causal structures, which will cause several problems for the parameter estimation in both linear modelling and linear graph learning. First, since linear model estimation is based on error minimization, multicollinearity will reduce the magnitude of the minimal eigenvalue in the linear space, causing numerical convergence problems (e.g., Cholesky decomposition or gradient descent) when applying maximum liklihood or maximum a posteriori. Second, severe multicollinearity amplifies parameter estimate instability across samples, making it difficult to interpret the coefficients reliably and accurately. Third, multicollinearity causes problems for statistical tests that rely on the sample covariance (e.g., the post-OLS t-test or the lasso covariance test \citep{lockhart2014significance}. The conditional correlation tests of in constraint-based learning \citep{farrar1967multicollinearity}). Last but not least, multicollinearity may also reduce the algorithmic stability of the model \citep{elisseeff2003leave}, which reduces the generalization ability and the out-of-sample prediction of the estimated model.

Moreover, multicollinearity also affects the reliability of variable selection algorithms in linear modelling and linear graph learning. \citet{zou2005regularization, jia2010model} find that, if a group of variables are highly correlated with one another, lasso-type estimators may randomly select one variable from the group and drop the others out of the regression, referred to as \textbf{the grouping effect}. Since all linear modelling techniques make the variable selection decision based on the conditional correlation between $\mathbf{x}_j$ and $\mathbf{y}$, the grouping effect may well apply to all variable selection methods in linear models. Multicollinearity also affect linear graph learning. \citet{heckerman95, chickering04} shows that learning a linear graph is NP-hard on data with large $p$. As a result, graph learning algorithms typically work well on data with large $n$ and very sparse $p$. In many graph learning applications, variable selection algorithms (e.g., SCAD \citep{fan2001variable}, ISIS \citep{fan2008sure} or different lasso-type estimators \citep{fan2009network}) are used to filter out redundant variables before graph learning. As a result, with grouping effect, variable selection methods may randomly drop some of the highly correlated variables, resulting in omissions of important variables in the linear graph learning.

Many attempts have been made to reduce the effects of multicollinearity. For more stable regression coefficient estimates, \citet{hoerlkennard70} apply Tikhonov regularization to OLS, resulting in the ridge regression. However, it complicates statistical tests and post-estimation inference. To reduce the grouping effect and obtain stable variable-selection results, cross-validated group lasso and cross-validated elastic net (CV-en) have been introduced \citep{zou2005regularization, friedman10}. However, group lasso relies on manual grouping of variables, which depends heavily on accurate field knowledge. On the other hand, while \citet{zou2005regularization} and \citet{jia2010model} show that CV-en may improve the stability of variable selection, \citet{jia2010model} counter that the improvement is marginal and that ``when the lasso does not select the true model, it is more likely that the elastic net does not select the true model either.''

\subsection{Main results}

In this paper we combine two well-known tools from machine learning and biostatistics---variable selection algorithms and graph learning---and apply them to estimate the causal structure of the housing market and the follow-up socio-economic effects using data for 2010 from Sydney, Australia. It is an ultrahigh dimensional database consisting of local education data, GIS information, census data, house characteristics and other socio-economic records. We show that, with "big data", it is possible to perform a data-driven instrument selection efficiently and purge out the invalid instruments. The estimated graph of the causal structure of the housing market provides an intuitive interpretation and matches the facts of the Sydney house market, economic theories and the previous empirical findings on house pricing. The estimated graph also returns an accurate and sparse house pricing model, outmatching other methods in terms of the bias-variance trade-off.

The estimated graph visually depicts the causal structure of house pricing dynamics. Using the graph, we detect endogeneity in house prices, which is confirmed by simultaneous equations modelling. The graph estimation method therefore represents a data-driven, as opposed to ad hoc, approach to detecting endogeneity. Furthermore, we are able to use the graph effectively and efficiently for instrument selection and validation, which are confirmed by traditional instrument tests from Durbin, Wooldridge and Hausman. Moreover, using the graph-recommended instrument, we significantly resolve endogeneity bias in the house price regression, which is confirmed by two-stage least squares. Last but not least, the graph estimation method also helps to identify a weak instrument, which is consistent with economic intuition.

The paper is organized as follows. In section~\ref{section:intro}, we introduce variable and instrument selection from the perspective of the random graph. In section~\ref{section:estimation}, we introduce the 2010 Sydney house data and demonstrate in detail the procedure of variable selection and graph estimation using the data. In section~\ref{section:application}, we use the estimation results for endogeneity detection and instrument selection. We also show that our graph-based results are consistent with received empirical knowledge on the housing market.

%%%%%%%%%%%%%%%%%%%%%%%%%%%%%%%%%%%%%%%
%%%%%%%%%%%% House data %%%%%%%%%%%%%%%
%%%%%%%%%%%%%%%%%%%%%%%%%%%%%%%%%%%%%%%

\section{Graph and instrument variable selection\label{section:intro}}

Before applying graphs for instrument variable selection and causal inference, it is important to introduce graphs and the graphical definition of instrument variables properly. Graph learning terminologies are defined differently across different areas. To consistent with the literature in machine learning, the following definitions and explanations on graph and instrument variables are based on \citet{spirtes2000causation}, \citet{pearl2009causality} and \citet{koller2009probabilistic}.

\subsection{Graphical criteria for instrument variables}

To properly define an instrument variable using graphs, we need first to define how the change in one variable can affect another in a graph. This is represented by the concept of a \textbf{trail}.\footnote{This is also referred to as ``path'' in some graph learning literature}

\begin{definition}[Trail of a graph]
  \label{def:trail}
\end{definition}
\begin{itemize}
  \item for any pair of variables $\left( \mathbf{x}_i, \mathbf{x}_j \right)$ in a graph, we say that they are \textbf{connected} ($\mathbf{x}_i \rightleftharpoons \mathbf{x}_j$) if either $\mathbf{x}_i \rightarrow \mathbf{x}_j$ or $\mathbf{x}_j \rightarrow \mathbf{x}_i$ ($\mathbf{x}_i$ and $\mathbf{x}_j$ have a parent-child relation).
  \item for variables $\mathbf{x}_1, \ldots, \mathbf{x}_k$ in a graph, we say that they form a \textbf{trail} if, $\forall\, 1 \leqslant i \leqslant k-1$, $\mathbf{x}_i \rightleftharpoons \mathbf{x}_{i+1}$.
\end{itemize}

Intuitively, a trail is a sequence of variables that are sequentially connected by arrows. A change in $\mathbf{x}_1$ can affect $\mathbf{x}_k$ only if there is a trail between the two variables. Put it from the perspective of joint distribution, if there exists a trail between $\mathbf{u}$ and $\mathbf{v}$, either the unconditional or some conditional correlation between $\mathbf{u}$ and $\mathbf{v}$ is not zero in population. In Figure~\ref{fig:example_dag}, for example, $\mathbf{z} \rightarrow \mathbf{x} \rightarrow \mathbf{y}$ is a trail, meaning a change in $\mathbf{z}$ can be passed to $\mathbf{y}$ if $\mathbf{x}$ is not conditioned on; or, equivalently, the correlation between $\mathbf{z}$ and $\mathbf{y}$ is not zero if we do not condition on $\mathbf{x}$. In Figure~\ref{fig:example_dag} and (\ref{eqn:instrument}), $\mathbf{z} \rightarrow \mathbf{x} \rightarrow \mathbf{y} \leftarrow \mathbf{u}$ is also a trail, meaning a change in $\mathbf{z}$ can pass to $\mathbf{u}$ only if (i) $\mathbf{y}$ is held constant and (ii) $\mathbf{x}$ is not fixed; from the perspective of correlation, $\mathbf{z}$ and $u$ are correlated when $\mathbf{y}$ is held constant and $\mathbf{x}$ is not.

In the examples above, $\mathbf{x}$ plays a key role in the trails. If $\mathbf{x}$ is held constant, the population correlation between $\mathbf{y}$ and $\mathbf{z}$ will be zero. As a result, any change in a variable at one end of the trail cannot affect the variable on the other end, which is the mistake that we make in example 2. To describe the role of variables like $\mathbf{x}$, we say the variables at both ends of the trail are \textbf{d-separated} by $\mathbf{x}$.\footnote{This is also referred to as ``the variables at both ends of the trail are \textbf{blocked} by $\mathbf{x}$'' in some graph learning literature}

\begin{definition}[d-separation]
  \label{def:d_separation}
  Let $P$ be a trail from the variable $u$ to the variable $v$. We define $u$ and $v$ to be d-separated by a set of variables $Z$ (denoted $u \ind v$ by $Z$) if $u$ and $v$ are independent after conditioning on all variables in $Z$.

  For example, $u \ind v$ by $Z$ in the following cases.
  \begin{itemize}
    \item $P$ contains a \textbf{directed chain} ($u \leftarrow \cdots \leftarrow m \leftarrow \cdots \leftarrow v$ or $u \rightarrow \cdots \rightarrow m \rightarrow \cdots \rightarrow v$) such that the middle variable $m \in Z$;
    \item $P$ contains a \textbf{fork} ($u \leftarrow \cdots \leftarrow m \rightarrow \cdots \rightarrow v$) such that the middle variable $m \in Z$;
    \item $P$ contains a \textbf{collider} ($u \rightarrow \cdots \rightarrow m \leftarrow \cdots \leftarrow v$) such that the middle variable $m \not\in Z$ and no descendant of $m$ is in $Z$.
  \end{itemize}
\end{definition}
\begin{figure}[H]
  \centering
  \includegraphics[width=0.2\paperwidth]{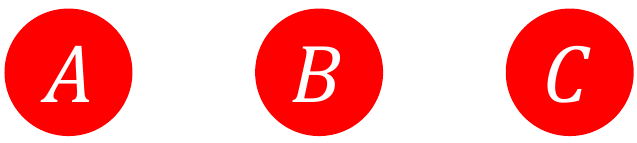}
  \caption{There does not exist any kind of dependency among $A$, $B$ and $C$. Hence, $A$ and $B$ are d-separated by any possible variable.}
  \label{fig:graphical_criteria_instrument}
\end{figure}

\begin{figure}[h]
  \centering
  \subfloat[\label{fig:instrument1}graph $G$, where $\mathbf{z} \not\ind \mathbf{x}$]
  {\includegraphics[width=0.2\paperwidth]{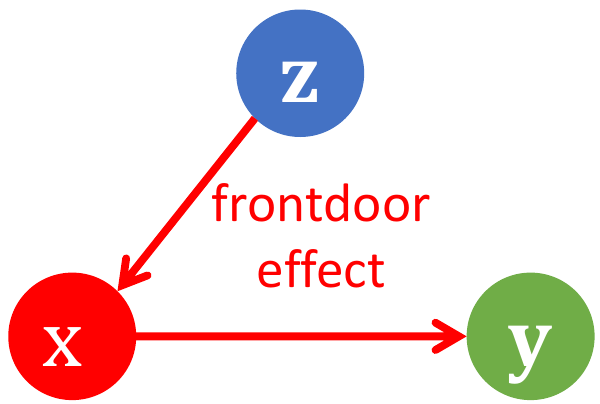}}
  \hfil
  \subfloat[\label{fig:instrument2}graph $G_{\overline{X}}$, where $\mathbf{z} \ind Y$]
  {\includegraphics[width=0.2\paperwidth]{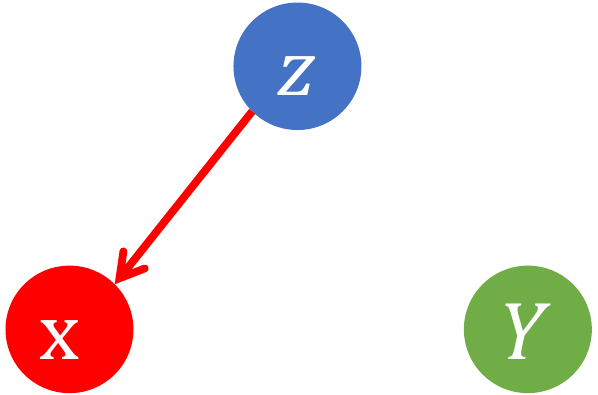}}
  \caption{Illustration: definition of an instrumental variable.}
  \label{fig:instrument}
\end{figure}

\noindent
We also introduce two useful remarks for d-separation. Firstly, if $A$ directly causes $B$ (i.e., $A \rightarrow B$) with no intermediate variables, $A$ and $B$ will never be independent regardless of the variable conditioned on (except $A$ and $B$). In this case, we say that \textit{no} variable can d-separate $A$ and $B$ (sometimes denoted $A \not\ind B$). Secondly, as illustrated in Figure~\ref{fig:graphical_criteria_instrument}, if $A$ and $B$ have no causal relation whatsoever, we say \textit{any} variable (for example, variable $C$) can d-separate $A$ and $B$ (sometimes denoted $A \ind B$). Using the concept of d-separation, the graphical definition of an instrument can be precisely defined, following \citet{brito2002generalized}, \citet{pearl2009causality} and \citet{silva2017learning}, and illustrated in Figure~\ref{fig:instrument}.\footnote{As shown by \citet{brito2002generalized} and \citet[pp.~247-248]{pearl2009causality}, the complete set of graphical criteria for an instrument is more complicated than our definition as it incorporates the idea of conditional instruments in a graph. To avoid being sidetracked, we leave further discussion to Appendix~\ref{App:IV_def}.}

\begin{definition}[Graphical criteria of instruments]
  \label{def:instrument_variable}
  Let $\mathbf{x}, \mathbf{z}$ and $\mathbf{y}$ be variables in graph $G$ and $\mathbf{x}$ directly causes $\mathbf{y}$. $\mathbf{z}$ is an instrument for $\mathbf{x}$ if
  \begin{itemize}
    \item[\textbf{G1}] $\mathbf{z}$ and $\mathbf{y}$ can be d-separated by any variable in $G_{\overline{\mathbf{x}}}$, where $G_{\overline{\mathbf{x}}}$ is the graph in which the effect from $\mathbf{x}$ to $\mathbf{y}$ is cut off (sometimes denoted $\left(\mathbf{z} \ind Y \right)_{G_{\overline{\mathbf{x}}}}$).
    \item[\textbf{G2}] $\mathbf{z}$ and $\mathbf{y}$ cannot be d-separated by any variable in $G$ (sometimes denoted $\left(\mathbf{z} \not\ind \mathbf{x} \right)_G$).
  \end{itemize}
\end{definition}

\noindent
Graphically, G1 means that, if we remove all the causal effects from $\mathbf{x}$ to $\mathbf{y}$, $\mathbf{z}$ cannot affect $\mathbf{y}$ any more.\footnote{Sometimes, G1 is modified to $\left(\mathbf{z} \ind Y \right)_{G_{\overline{\mathbf{x}}}}$, where $G_{\overline{\mathbf{x}}}$ is obtained by removing all the arrows entering $\mathbf{x}$ from the graph $G$, but Definition~~\ref{def:instrument_variable} is the more common definition of instrument. Nonetheless, both versions mean that the effect from $\mathbf{z}$ to $\mathbf{y}$ must go only through $\mathbf{x}$.} Similarly, G2 means that the effect from $\mathbf{z}$ to $\mathbf{x}$ cannot be broken by holding any variable constant. Both G1 and G2 mean that the effect from $\mathbf{z}$ to $\mathbf{y}$ must go only through $\mathbf{x}$. Put another way, holding $\mathbf{x}$ constant, $\mathbf{z}$ cannot affect $\mathbf{y}$ by any means. In graph learning, the effect from $\mathbf{z}$ to $\mathbf{y}$ via the (endogenous) variable $\mathbf{x}$ is also referred to as the \textbf{frontdoor effect (FE)} (Figure~\ref{fig:instrument1}). Moreover, Definition~\ref{def:instrument_variable} is a generalized version of the usual definition of an instrument in regression analysis. Assuming that $\mathbf{x}$ causes $\mathbf{y}$ in (\ref{eqn:instrument}), $\mathrm{corr} \left( \mathbf{z}, \mathbf{x} \right) \neq 0$ implies the existence of an FE. Likewise, $\mathrm{corr} \left( \mathbf{z}, u \right) = 0$ implies there does not exist any effect from $\mathbf{z}$ to $\mathbf{y}$ that does not go through $\mathbf{x}$ (also referred to as \textbf{no backdoor effect (BE)}).\footnote{In other graph learning literature (\citet{pearl2009causality}, for example), this is also referred to as ``$\mathbf{x}$ satistfies the `no backdoor criterion' for the causal effect between $\mathbf{z}$ and $\mathbf{y}$''}

The classical machine learning and biostatistics researches (for example, \citet{spirtes2000causation} and \citet{pearl2009causality}) show that definition~\ref{def:instrument_variable} (aka the graphical criteria of instrument variables) implies the classical definition based on regression error, which is also referred to as the error-based criteria of instrument variables. For more detailed analysis and examples, see Appendix~\ref{App:IV_def}.

\begin{figure}[H]
	\centering
	\includegraphics[width=0.15\paperwidth]{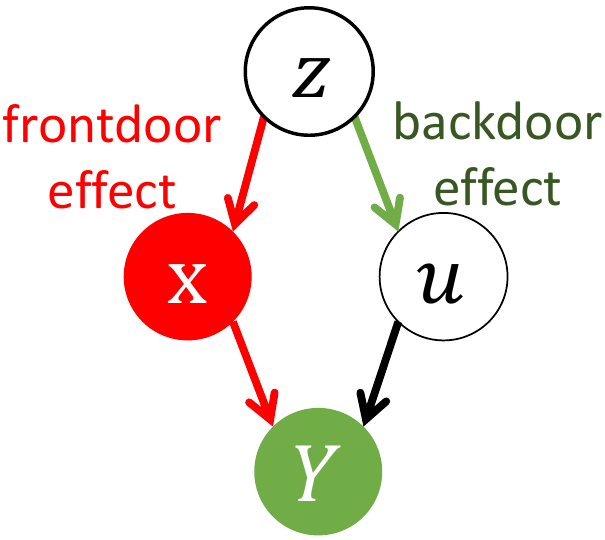}
	\caption{Illustration: violation of Definition~\ref{def:instrument_variable}.}
	\label{fig:not_instrument1}
\end{figure}

\noindent
As \citet{spirtes2000causation} and \citet{pearl2009causality} show, Definition~\ref{def:instrument_variable} can be used to identify instruments variables in a graph. Take Figure~\ref{fig:not_instrument1} as an example. A classical case in econometrics, Figure~\ref{fig:not_instrument1} contains an arrow from $\mathbf{z}$ to $u$. As a result, $\mathrm{corr} \left( \mathbf{z}, u \right) \neq 0$ in (\ref{eqn:instrument}), implying $\mathbf{z}$ is not a valid instrument. Equivalently, the arrow from $\mathbf{z}$ to $u$ allows $\mathbf{z}$ to affect $\mathbf{y}$ separately from the endogenous variable $\mathbf{x}$, which is a BE. As a result, Figure~\ref{fig:not_instrument1} violates G1 since $\mathbf{z}$ and $\mathbf{y}$ are not independent even though $\mathbf{x}$ is held constant.

\subsection{Variable selection for graph learning}

Figure~\ref{fig:instrument} and Appendix~\ref{App:IV_def} reveals the unparalleled advantage of the graph on instrument variable selection. If we can accurately estimate the graph (or at least estimate the role of each variable relative to $\mathbf{y}$), choosing an appropriate instrumental variable is straightforward. Hence, an accurate graph estimation is the core of data-drive causal inference. As example 2 and 3 illustrate, variable omission and variable redundancy can both mislead the graph learning and, consequently, causal inference. To avoid variable omission, it is always safe to start graph learning from "big data" --- a dataset that contains as many relevant variables as possible. Unfortunately, minimizing the chance of variable omission may bring a large number of potential variables in your graph, raising the issues of dimensionality and computation. Dimensionality and computation aside, including redundant variables may cause problems in example 2 and 3 and return misleading and counterintuitive results. Hence, it is necessary to accompany graph learning with variable selection (aka variable elimination).

To build the connection between variable selection and regression-based causal inference, in this paper we follow the common linearity assumptions (e.g., \citet{bollen1989structural, geiger1994learning, spirtes2000causation}) as follows,
\begin{itemize}
  \item[\textbf{A1}] the data generating process of each variable can be represented as a linear regression equation;
  \item[\textbf{A2}] the dependencies among variables can be represented by correlation (e.g., (\ref{eqn:instrument})).
\end{itemize}
Both \textbf{A1} and \textbf{A2} imply that the causal structure is linear or, equivalently, we have a linear graph in population. It is worth noting that we do not assume that the linear equation is a perfect representation of the data generating process. In fact, it is quite common for linear models to suffer misspecification, especially if we are not sure about the linearity of the data-generating process. Hence, in this paper, we start graph learning from linear graph and always check the appropriateness of linearity assumption when we get the result. If linearity is not appropriate in some case, we will add nonlinearity pattern (e.g., neural network, kernel regression in reproducing kernel Hilbert spaces) into the graph learning.

\begin{figure}[h]
  \centering
  \includegraphics[width=0.3\paperwidth]{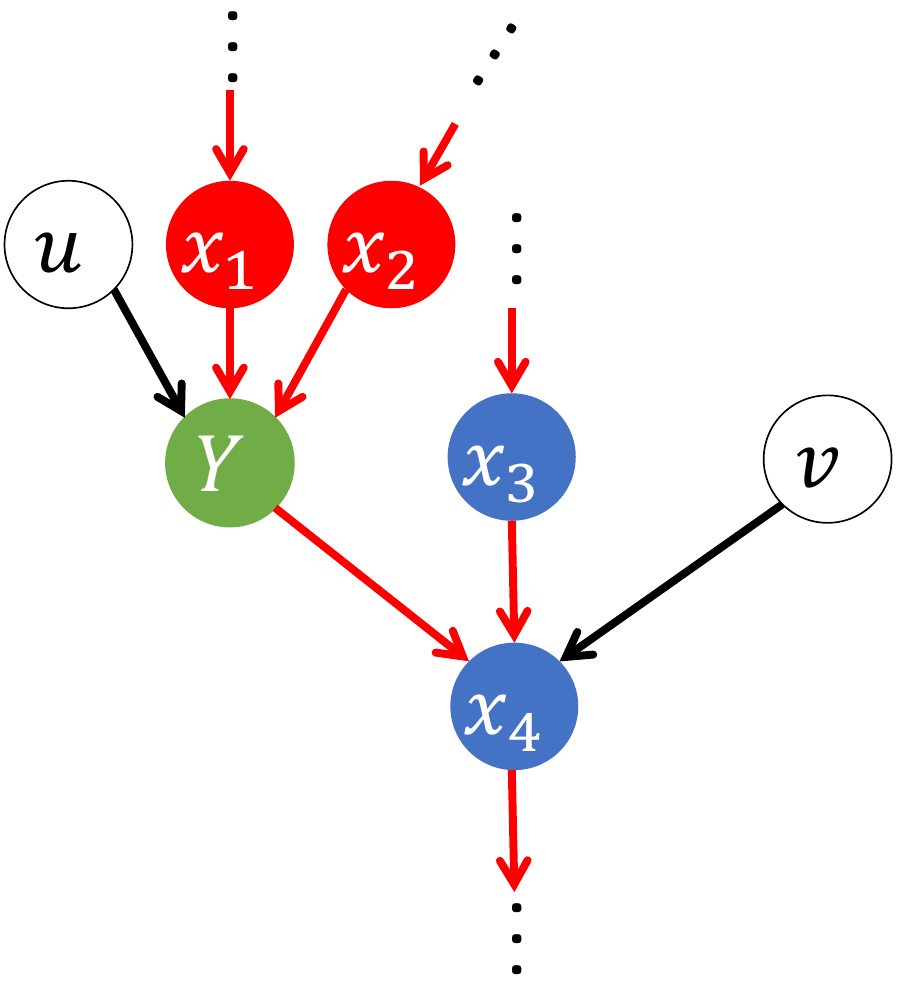}
  \caption{Illustration: recovering the Markov blanket for $\mathbf{y}$.}
  \label{fig:variable_selection}
\end{figure}

With \textbf{A1} and \textbf{A2}, all graphs in this paper are linear graphs and graph learning can be comprehended from the perspective of high-dimensional regression analysis. In classical regression analysis, significance test and variable selection algorithms are applied to find the variables with non-zero population coefficients. A regression coefficient represents the conditional correlation between the corresponding covariate and the response variable, holding other covariates constant. As a result, variable selection algorithms aim to find the variables that are conditionally correlated to $\mathbf{y}$ in the population, holding all other variables constant.\footnote{After standardizing the response variable and all covariates, the regression coefficient of $\mathbf{x}_i$ is the conditional correlation between $\mathbf{x}_i$ and $\mathbf{y}$, holding all other covariates constant.} In graph learning, the set of such variable is called \textbf{Markov blanket of $\mathbf{y}$} (denoted MB($\mathbf{y}$)), which includes the parent(s), children and spouse(s) of $\mathbf{y}$. Hence, in linear graphs, recovering the MB($\mathbf{y}$) is equivalent to finding true variables in the linear regression of $\mathbf{y}$ on all other variables, illustrated graphically in Figure~\ref{fig:variable_selection} and analysed with example 4.\footnote{For a more general explanation and examples, see \citet{pearl2009causality}, \citet{koller2009probabilistic} or \citet{scutari2014bayesian}.}

\noindent
\textbf{Example 4.} In Figure~\ref{fig:variable_selection}, $\mathbf{u}$ and $\mathbf{v}$ are independent latent noise terms; $\left\{\mathbf{x}_1, \mathbf{x}_2, \mathbf{u}\right\}$ are the parents of $\mathbf{y}$; $\left\{\mathbf{x}_3, \mathbf{v}\right\}$ are the spouses of $\mathbf{y}$; $\left\{\mathbf{y}, \mathbf{x}_3, \mathbf{v}\right\}$ together cause $\mathbf{x}_4$. Together with A1 and A2, the data-generating process in Figure~\ref{fig:variable_selection} is the following linear regression system,
\begin{equation}
  \begin{cases}
    \mathbf{y}   & = \alpha_0 + \alpha_1 \mathbf{x}_1 + \alpha_2 \mathbf{x}_2 + \mathbf{u},\\
    \mathbf{x}_4 & = \beta_0  + \beta_1 \mathbf{y}    + \beta_1 \mathbf{x}_3  + \mathbf{v}.
  \end{cases}
  \label{eqn:variable_selection}
\end{equation}
Holding $\left\{\mathbf{x}_1, \mathbf{x}_2\right\}$ constant, (\ref{eqn:variable_selection}) shows that all the variation in $\mathbf{y}$ is caused only by $\mathbf{u}$ (mathematically, $\mathbf{y} \vert \left\{\mathbf{x}_1, \mathbf{x}_2\right\} = \mathbf{u}$). Put another way, after partialing out $\left\{ \mathbf{x}_1, \mathbf{x}_2 \right\}$ from $\mathbf{y}$, the variation in $\mathbf{u}$ can be explained by the children of $\mathbf{y}$. As a result, the independence between $\mathbf{v}$ and $\left\{ \mathbf{x}_1, \mathbf{x}_2 \right\}$ and the second equation of (\ref{eqn:variable_selection}) imply that
\begin{align}
  \mathbf{u} = & \mathbf{y} \vert \left\{\mathbf{x}_1, \mathbf{x}_2\right\} \notag \\
    = & - \frac{\beta_0}{\beta_1}  + \frac{1}{\beta_1} \mathbf{x}_4 \vert \left\{\mathbf{x}_1, \mathbf{x}_2\right\} - \frac{\beta_2}{\beta_1} \mathbf{x}_3 \vert \left\{\mathbf{x}_1, \mathbf{x}_2\right\} - \frac{\mathbf{v}}{\beta_1} .
  \label{eqn:variable_selection_example}
\end{align}
After replacing $\mathbf{u}$ in (\ref{eqn:variable_selection}) with the right-hand side of (\ref{eqn:variable_selection_example}), the data generating process of $\mathbf{y}$ --- first equation in (\ref{eqn:variable_selection}) --- reduces to the following population regression equation of $\mathbf{y}$ on its MB members,
\begin{align}
  \mathbf{y} = \gamma_0 + \gamma_1 \mathbf{x}_1 + \gamma_2 \mathbf{x}_2 + \gamma_3 \mathbf{x}_3 + \gamma_4 \mathbf{x}_4 + \mathbf{e}.
  \label{eqn:reduced_form_example}
\end{align}
%
%#DONE: explain the endogeneity between e and  children of Y
%
where $\mathbf{e}$ is a linear function of $\mathbf{v}$. Hence, Equation~(\ref{eqn:reduced_form_example}) is the population reduced form of the linear system (\ref{eqn:variable_selection}), where only MB variables are informative (or true) variables.$\qed$

\bigskip
Example 4 means that, when we apply the variable selection algorithm on Equation~(\ref{eqn:reduced_form_example}) with enough sample size, a consistent variable selector (for example, lasso-type estimators and SCAD) should keep only $\left\{ \mathbf{x}_1, \mathbf{x}_2, \mathbf{x}_3, \mathbf{x}_4 \right\}$ and purge out all other variables. \citet{zhaoyu06} show that the irrepresentable condition almost surely guarantees the variable selection consistency of lasso. This implies that, with irrepresentable condition, it is very likely that lasso only select the true variables in Equation~(\ref{eqn:reduced_form_example}) --- the MB members of $\mathbf{y}$ --- with small enough $p/n$. As a result, in linear graphs, variable selection for the regression of $\mathbf{y}$ is equivalent to finding the MB of $\mathbf{y}$. Having said that, due to the complicated causal structure, variables in the graph may be heavily correlated. As a result, when applying lasso-type estimators in a graph, we need to always check correlation between the selected variables and dropped variables in order to detect potential violation of the irrepresentable condition.

Finding the correct Markov Blanket of a variable is important for graph learning. In graph learning, both score-based and constraint-based learning works combinatorially, which require huge computation load. As a result, the classical graph learning algorithm do not work well when dimensionality is high. However, after purging as many redundant variables as possible from the Markov blanket, the possible combination number of causal effects from and to $\mathbf{y}$ will be reduced exponentially. As a result, accurate Markov Blanket estimation makes graph learning easy and quick, which further faciliates causal inference and instrument variable selection on $\mathbf{y}$.

\section{Graph estimation on Sydney house market data \label{section:estimation}}

In this section, we prepare the Sydney real estate database for graph learning and causal inference. In the classical applied econometrics, the house price is typically explained by easily-measured attributes (e.g., the number of bedrooms, bathrooms, land size, distance to amenities, etc.) using a linear regression equation. To avoid the possible variable omission bias, our database includes as much objective information that is relevant to the market value of particular a house, much of which, of course, is determined by the location of the property, its unique features, and the characteristics of the neighbourhood. With a "big data" database, we apply the variable selection algorihtm to the database and try to find as many Markove blanket memebers of house price as possible.

\subsection{Description and sources of databases and primitive variable elimination}

The house market database is assembled from more than 10 different datasets, including 2010 Sydney house transaction data (including every 2010 sale of a house in City, Mid, North, South and East Sydney as well as all house features), 2010 and 2011 Sydney crime data by suburb, 2010 GIS data (extracted and complied from Sydney geospatial topological database, climate database, pollution database and Google Maps database), 2011 census data by Statistical Area Level 1 (SA1, the smallest census area in Australia, which have an average population size of approximately 200 people or, equivalently, 60 households), 2009 local school quality and catchment data, 2010 Sydney traffic data, data on public transport (train stations, ferry docks and bus routes), and so on. To speed up computation and data manipulation, we synthesize all databases using \textsf{Aparche Spark} on the Google Cloud. Altogether, the total variable number is above 10 thousands, the observations number is above 10 thousands and the size of the database is above 200GB. Due to the size of the synthesized database, we do not attach it in this paper.

The variables in the databases are collected from different sources. Some house features are reported in real-estate advertising and others are scraped from Google searches using a \textsf{Python} internet scraper; the distance of each house to nearest key locations is computed in QGIS---a open-source \textsf{Python}-based geographical information system---using the GPS location of each house and geodata collected from Google Maps and Department of Land and Natural Resources, New South Wales. The 2009 Index of Community Socio-Educational Advantage (ICSEA) score---an measure of the socio-educational background of students at each school---is collected from the Australian Curriculum, Assessment and Reporting Authority (ACARA). The variables on local school quality (average National Assessment Program -- Literacy and Numeracy (NAPLAN) examination results) are also collected from ACARA. The 2009 and 2010 crime data are collected from the Australian Bureau of Statistics and Department of Justice, New South Wales. The 2011 census data, traffic data, climate data, geospatial topological database and pollution database  are acquired from the Australian Bureau of Statistics. It is also worth noting that all the socio-economic data are observed by SA1, the smallest statistic area in 2011 census. Each SA1 in our data contains typically around 200 local residents. Most important, to avoid Simpson's paradox, in QGIS we only incorporate the SA1 that only covers houses, which rule out other types of real estate like apartments.\footnote{In other graph learning literature, Simpson's paradoxis also referred to as Simpson's reversal or reversal paradox. For example, see \citet{simpson1951interpretation} and \citet{blyth1972simpson}.} The detailed variable list is attached at Appedix~2 as a csv file.

As explained in literature review, graph learning and MB selection typically work well on datasets with small $p/n$. However, our dataset has $n$ and $p$ both larger than 10000. Not only is that well above the computation limit of graph learning but, with such high dimensionality, there are likely to be many variables irrelevant to the MB of house price. As a result, following \citet{fan2008sure}, we conduct variable selection in two stages to reduce dimensionality from high to a moderate scale that is below the sample size. As the primitive stage, we use iterative sure independence screening (ISIS, \citet{fan2008sure}) and rule out variables whose conditional correlation to house prices are, ceteris paribus, approximately $0$. Using the variables that survive ISIS, we execute detailed variable selection and use the corresponding results for MB selection and graph learning.

To control the effect of high dimnesionality and sampling randomness, we embed ISIS into the framework of boostrap. The detailed step of ISIS is explained as follows. Firstly, we generate 2000 bootstrap samples. On each bootstrap sample, we run ISIS directly to select the variables that are highly correlated to house price. The ISIS is stopped based on BIC minimization. After obtaining 2000 ISIS results, we average them and select the variables that is at least selected in 70\% of 2000 results. The last step is, considering the huge multicollinearity among variables that may render the selection result unstable, we also include the variables that are moderately correlated with the variable selected in last step. For example, only the year 3 mean reading score and year 5 numeracy score are selected by ISIS. Considering the possible grouping effect among mean scores of local school, we include them all. After conducting the ISIS variable elimination directly on Google Cloud, the variables that survive ISIS are returned as the first column of Table~\ref{table:house_variable}. As shown in the table, the 57 variables that survive SIS fall into 5 categories: features of the house, distances to key locations (public transport, shopping, etc.),  neighbourhood socio-economic data, localized administrative and crime data, and local school quality. Pairwise correlations among all 57 covariates indicate that, not surprisingly, multicollinearity and the grouping effect are present in the data.\footnote{Due to the large number of covariates, we report the correlations in supplementary files.} Thus, we proceed to variable selection with the upmost caution.

\begin{table}
  \scriptsize
  \caption{Variable selection by CV-en, CV-lasso (lars and cd) and Solar for linear and log models in Sydney house price data \label{table:house_variable}}
  \begin{tabular}{@{}ll@{\extracolsep{6pt}}c@{\extracolsep{-2pt}}c@{\extracolsep{6pt}}c@{\extracolsep{-2pt}}c@{\extracolsep{6pt}}c@{\extracolsep{-2pt}}c@{}}
    \toprule
            &             & \multicolumn{2}{c}{CV-en}
                          & \multicolumn{2}{c}{CV-lasso}
                          & \multicolumn{2}{c}{solar} \\
            &             &
                          &
                          & \multicolumn{2}{c}{(lar, cd)}
                          & \\
                          \cline{3-4} \cline{5-6} \cline{7-8} \\[-7pt]
    Variable & Description& \multicolumn{1}{c}{linear}
                          & \multicolumn{1}{c}{log}
                          & \multicolumn{1}{c}{linear}
                          & \multicolumn{1}{c}{log}
                          & \multicolumn{1}{c}{linear}
                          & \multicolumn{1}{c}{log} \\
    \midrule
    Bedrooms           & property, number of bedrooms             & \checkmark  & \checkmark  & \checkmark  & \checkmark  & \checkmark & \checkmark  \\
    Baths              & property, number of bathrooms            & \checkmark  & \checkmark  & \checkmark  & \checkmark  & \checkmark & \checkmark  \\
    Parking            & property, number of parking spaces       & \checkmark  & \checkmark  & \checkmark  & \checkmark  & \checkmark & \checkmark  \\
    AreaSize           & property, land size                      & \checkmark  & \checkmark  & \checkmark  & \checkmark  &   &    \\ \midrule
    Airport            & distance, nearest airport                & \checkmark  & \checkmark  & \checkmark  & \checkmark  &   &    \\
    Beach              & distance, nearest beach                  & \checkmark  & \checkmark  & \checkmark  & \checkmark  & \checkmark & \checkmark  \\
    Boundary           & distance, nearest suburb boundary        & \checkmark  & \checkmark  & \checkmark  & \checkmark  &   &    \\ Cemetery           & distance, nearest cemetery               & \checkmark  & \checkmark  & \checkmark  &    &   &    \\
    Child care         & distance, nearest child-care centre      & \checkmark  & \checkmark  & \checkmark  & \checkmark  &   & \checkmark  \\
    Club               & distance, nearest club                   & \checkmark  & \checkmark  & \checkmark  & \checkmark  &   &    \\
    Community facility & distance, nearest community facility     & \checkmark  & \checkmark  &    &    &   &    \\
    Gaol               & distance, nearest gaol                   & \checkmark  & \checkmark  &    &    & \checkmark & \checkmark  \\
    Golf course        & distance, nearest golf course            & \checkmark  & \checkmark  & \checkmark  & \checkmark  &   &    \\
    High               & distance, nearest high school            & \checkmark  & \checkmark  & \checkmark  & \checkmark  &   &    \\
    Hospital           & distance, nearest general hospital       & \checkmark  & \checkmark  &    & \checkmark  &   &    \\
    Library            & distance, nearest library                & \checkmark  &    & \checkmark  &    &   &    \\
    Medical            & distance, nearest medical centre         & \checkmark  & \checkmark  &    & \checkmark  &   &    \\
    Museum             & distance, nearest museum                 & \checkmark  & \checkmark  & \checkmark  & \checkmark  &   &    \\
    Park               & distance, nearest park                   & \checkmark  & \checkmark  & \checkmark  &    &   &    \\
    PO                 & distance, nearest post office            & \checkmark  & \checkmark  &    & \checkmark  &   &    \\
    Police             & distance, nearest police station         & \checkmark  & \checkmark  & \checkmark  & \checkmark  &   &    \\
    Pre-school         & distance, nearest preschool              & \checkmark  & \checkmark  & \checkmark  & \checkmark  &   &    \\
    Primary            & distance, nearest primary school         & \checkmark  & \checkmark  & \checkmark  & \checkmark  &   &    \\
    Primary High       & distance, nearest primary-high school    & \checkmark  & \checkmark  & \checkmark  & \checkmark  &   &    \\
    Rubbish            & distance, nearest rubbish incinerator    & \checkmark  & \checkmark  & \checkmark  &    &   &    \\
    Sewage             & distance, nearest sewage treatment       & \checkmark  &    &    &    &   &    \\
    SportsCenter       & distance, nearest sports centre          & \checkmark  & \checkmark  & \checkmark  & \checkmark  &   &    \\
    SportsCourtField   & distance, nearest sports court/field     & \checkmark  & \checkmark  & \checkmark  & \checkmark  &   &    \\
    Station            & distance, nearest train station          & \checkmark  & \checkmark  & \checkmark  &    &   &    \\
    Swimming           & distance, nearest swimming pool          & \checkmark  & \checkmark  & \checkmark  & \checkmark  &   &    \\
    Tertiary           & distance, nearest tertiary school        & \checkmark  & \checkmark  & \checkmark  & \checkmark  &   &    \\
    \midrule
    Mortgage           & SA1, mean mortgage repayment (log)       & \checkmark  & \checkmark  & \checkmark  & \checkmark  & \checkmark & \checkmark  \\
    Rent               & SA1, mean rent (log)                     & \checkmark  & \checkmark  & \checkmark  & \checkmark  & \checkmark & \checkmark  \\
    Income             & SA1, mean family income (log)            & \checkmark  & \checkmark  & \checkmark  & \checkmark  & \checkmark & \checkmark  \\
    Income (personal)  & SA1, mean personal income (log)          & \checkmark  &    &    &    &   &    \\
    Household size     & SA1, mean household size                 & \checkmark  & \checkmark  & \checkmark  & \checkmark  &   &    \\
    Household density  & SA1, mean persons to bedroom ratio       & \checkmark  & \checkmark  & \checkmark  & \checkmark  &   &    \\
    Age                & SA1, mean age                            & \checkmark  & \checkmark  & \checkmark  & \checkmark  &   & \checkmark  \\
    English spoken     & SA1, percent English at home             & \checkmark  & \checkmark  & \checkmark  &    &   &    \\
    Australian born    & SA1, percent Australian-born             & \checkmark  & \checkmark  & \checkmark  &    &   &    \\
    \midrule
    Suburb area        & suburb, area                             & \checkmark  &    & \checkmark  & \checkmark  &   &    \\
    Population         & suburb, population                       & \checkmark  & \checkmark  &    & \checkmark  &   &    \\
    TVO2010            & suburb, total violent offences, 2010     & \checkmark  & \checkmark  &    &    &   &    \\
    TPO2010            & suburb, total property offences, 2010    & \checkmark  & \checkmark  &    & \checkmark  &   &    \\
    TVO2009            & suburb, total violent offences, 2009     & \checkmark  & \checkmark  & \checkmark  &    &   &    \\
    TPO2009            & suburb, total property offences, 2009    & \checkmark  & \checkmark  &    &    &   &    \\
    \midrule
    ICSEA              & local school, ICSEA                      & \checkmark  & \checkmark  & \checkmark  & \checkmark  & \checkmark & \checkmark  \\
    ReadingY3          & local school, year 3 mean reading score  & \checkmark  & \checkmark  & \checkmark  & \checkmark  &   &    \\
    WritingY3          & local school, year 3 mean writing score  & \checkmark  & \checkmark  & \checkmark  & \checkmark  &   &    \\
    SpellingY3         & local school, year 3 mean spelling score & \checkmark  & \checkmark  & \checkmark  &    &   &    \\
    GrammarY3          & local school, year 3 mean grammar score  & \checkmark  & \checkmark  & \checkmark  &    &   &    \\
    NumeracyY3         & local school, year 3 mean numeracy score & \checkmark  & \checkmark  & \checkmark  & \checkmark  &   &    \\
    ReadingY5          & local school, year 5 mean reading score  & \checkmark  &    &    &    &   &    \\
    WritingY5          & local school, year 5 mean writing score  & \checkmark  & \checkmark  & \checkmark  &    &   &    \\
    SpellingY5         & local school, year 5 mean spelling score & \checkmark  & \checkmark  & \checkmark  &    &   &    \\
    GrammarY5          & local school, year 5 mean grammar score  & \checkmark  & \checkmark  & \checkmark  &    &   &    \\
    NumeracyY5         & local school, year 5 mean numeracy score & \checkmark  & \checkmark  &    &    &   &    \\
    \midrule
                       & Number of variables selected             & 57 & 53 & 44 & 36 & 9 & 11 \\

    \bottomrule

  \end{tabular}

\end{table}

The database is particularly well suited to demonstrate the linear graph learning technique. Firstly, we have over 200GB of data on tens of thousands of variables and more than ten thousands observations. The size of such `big data' reduces the possibility of variable omission. Even though some factors in the house market are not observable, their proxies are likely to be included in the synthesized database, further reducing the issue of variable omission. Secondly, as shown below, the regression $R^2$ on the selected variables are high (e.g., for a typical $\log \left( \mathrm{price} \right)$ regression with only 11 variables, $R^2=73\%$; slightly tuning the functional forms, the regression $R^2$ can easily reach approximately $90\%$), indicating that the majority of the patterns in the data are linear. As a result, for this application, the nonlinearity issue does not concern us too much. By constrast, in other applications of linear graph learning it would be prudent to carry out similar checks for variable omission and linearity.

\subsection{Second stage variable selection, its sparsity and prediction accuracy under different functional forms}

\citet{pearl2009causality} points out that dependence and causation relations should not be affected by the functional forms of variables. For example, if $\mathbf{y}$ is a parent of $\mathbf{x}$, $\mathrm{log} \left( \mathbf{y} \right) \rightarrow \mathrm{log} \left( \mathbf{x} \right)$ must be also true, and vice versa. Nonetheless, to avoid being misled by variable form, we conduct second stage variable selection in both linear and log terms, only selecting variables that are simultaneously selected in both scenarios. To avoid the possibility that some variable selection algorithm loses sparsity or accuracy, we implement solar, lasso and cross-validated elastic net (CV-en) for comparison. We optimize lasso using both cross-validated coordinate descent (CV-cd) and cross-validated least-angle regression (CV-lars), both of which return the same variable selection result due to $p/n \leqslant 1/200$.

With all variables in linear form, Table~\ref{table:house_variable} shows the selection results from solar, lasso and CV-en. Consistent with \citet{jia2010model}, both lasso solvers and CV-en simultaneously lose sparsity of variable selection due to the complicated causal structures and severe multicollinearity in the data. Lasso only manages to drop 7 variables and CV-en selects all 57 variables. It is not recommended to heuristically increase the value of $\lambda$ in lasso-type estimators (e.g., the one-sd rule or the `elbow' rule) since it may trigger further grouping effects and consequently lead to the random dropping of variables. On the other hand, CV-en is designed to tolerate multicollinearity and the grouping effect and is expected to return a sparse and stable regression result. However, CV-en fails to accomplish any variable selection, suggesting sensitivity to the complicated causal structure in the house price dataset. By contrast, solar returns a sparse regression model, with only $9$ variables selected from $57$.

\begin{figure}[H]
  \centering
  \subfloat[\label{fig:Hist_example_Beach}Beach.]
  {\includegraphics[width=0.22\paperwidth]{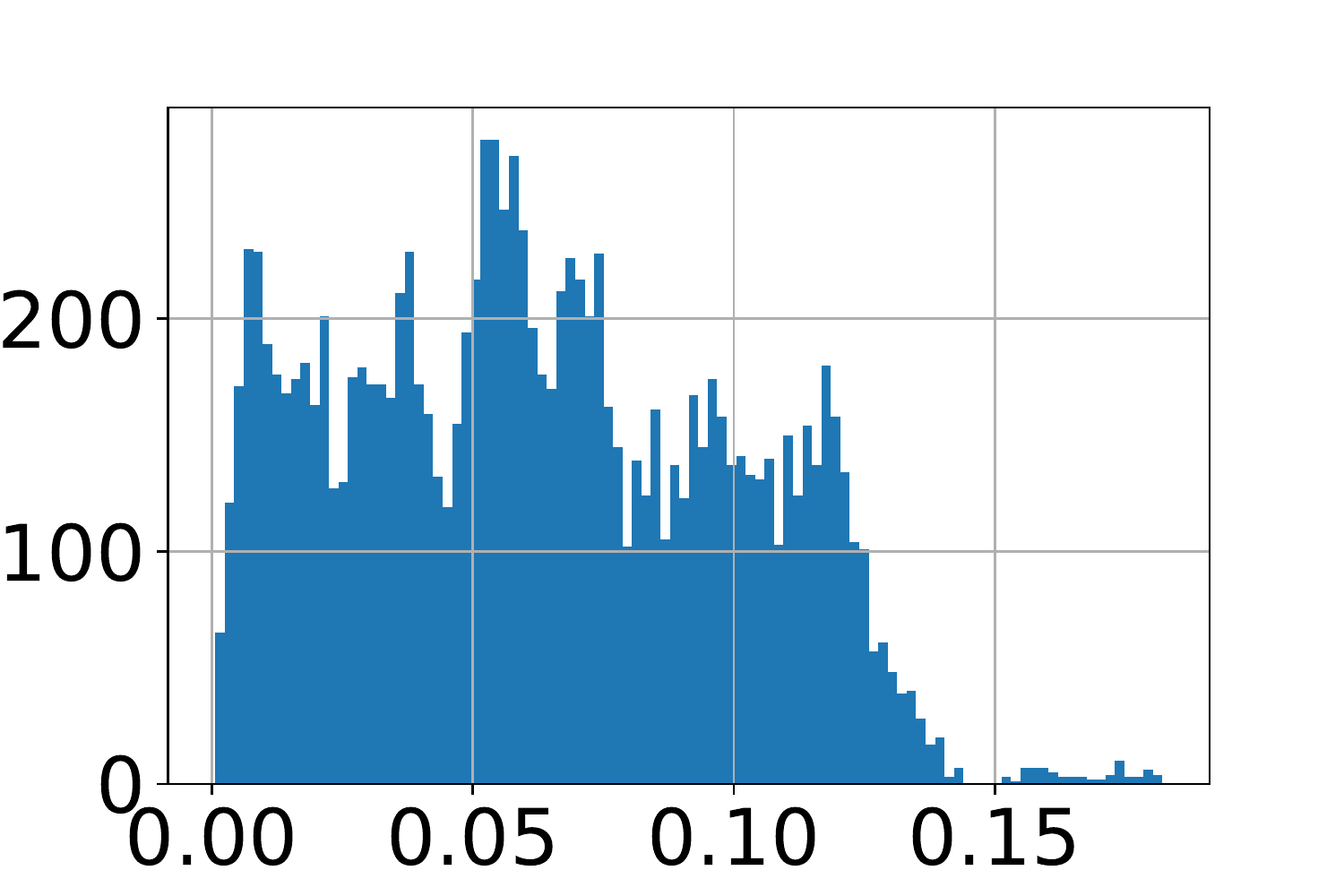}}
  \subfloat[\label{fig:Hist_example_logBeach}logBeach.]
  {\includegraphics[width=0.22\paperwidth]{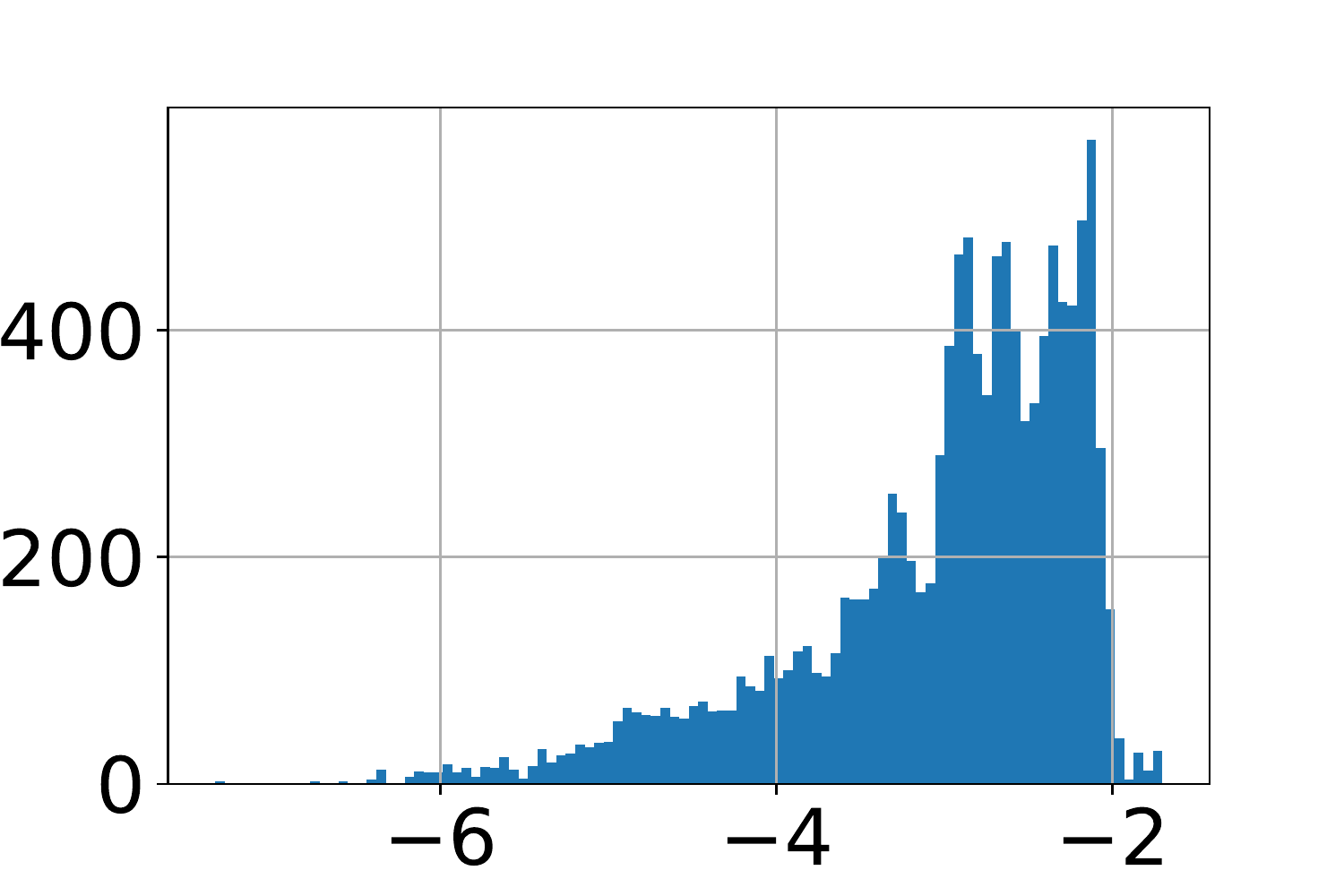}}
  \subfloat[\label{fig:Hist_example_logBeach1000}$\mathrm{log} \left( \mathrm{Beach} \cdot 1000 \right)$.]
  {\includegraphics[width=0.22\paperwidth]{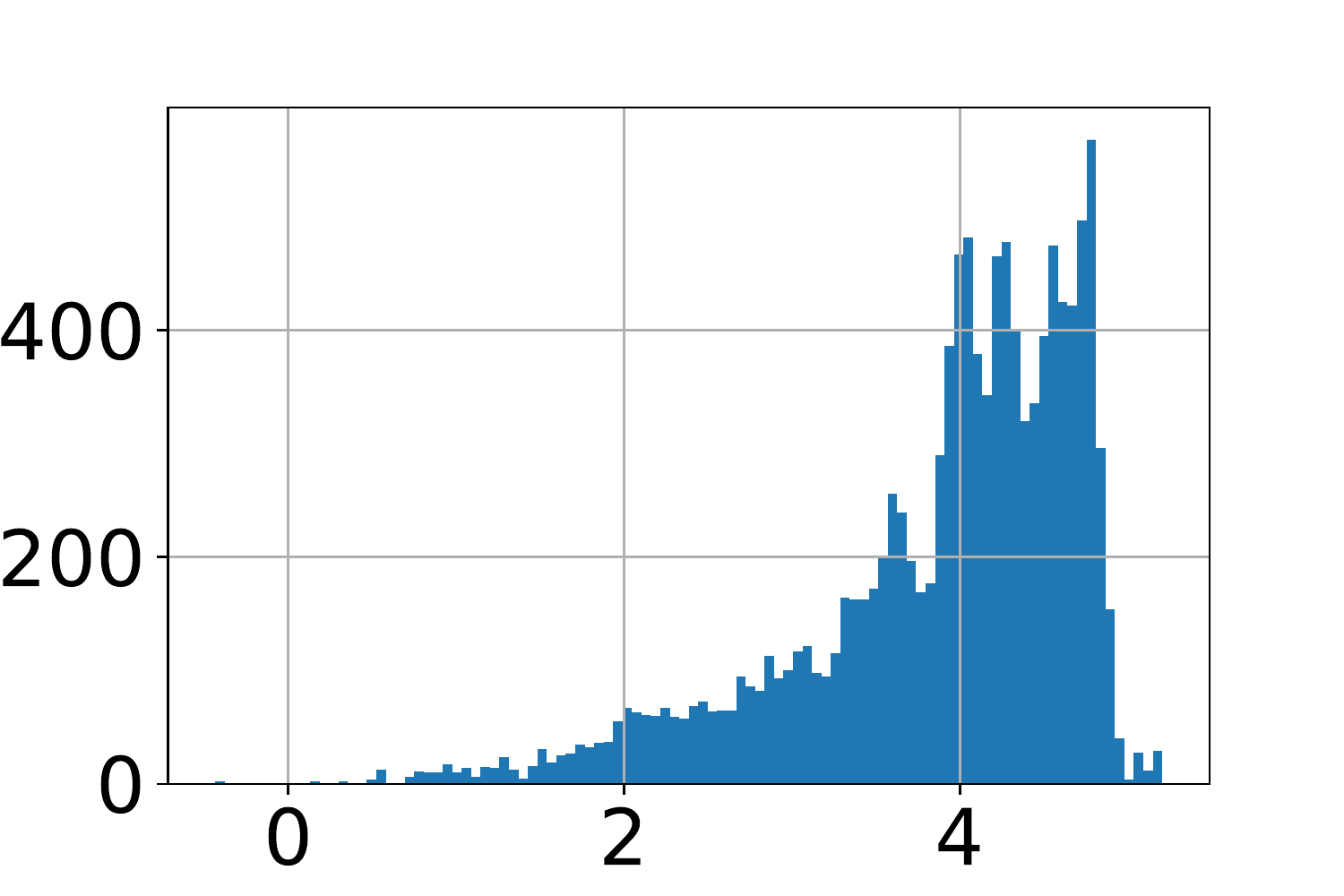}}

  \subfloat[\label{fig:Hist_example_Gaol}Gaol]
  {\includegraphics[width=0.22\paperwidth]{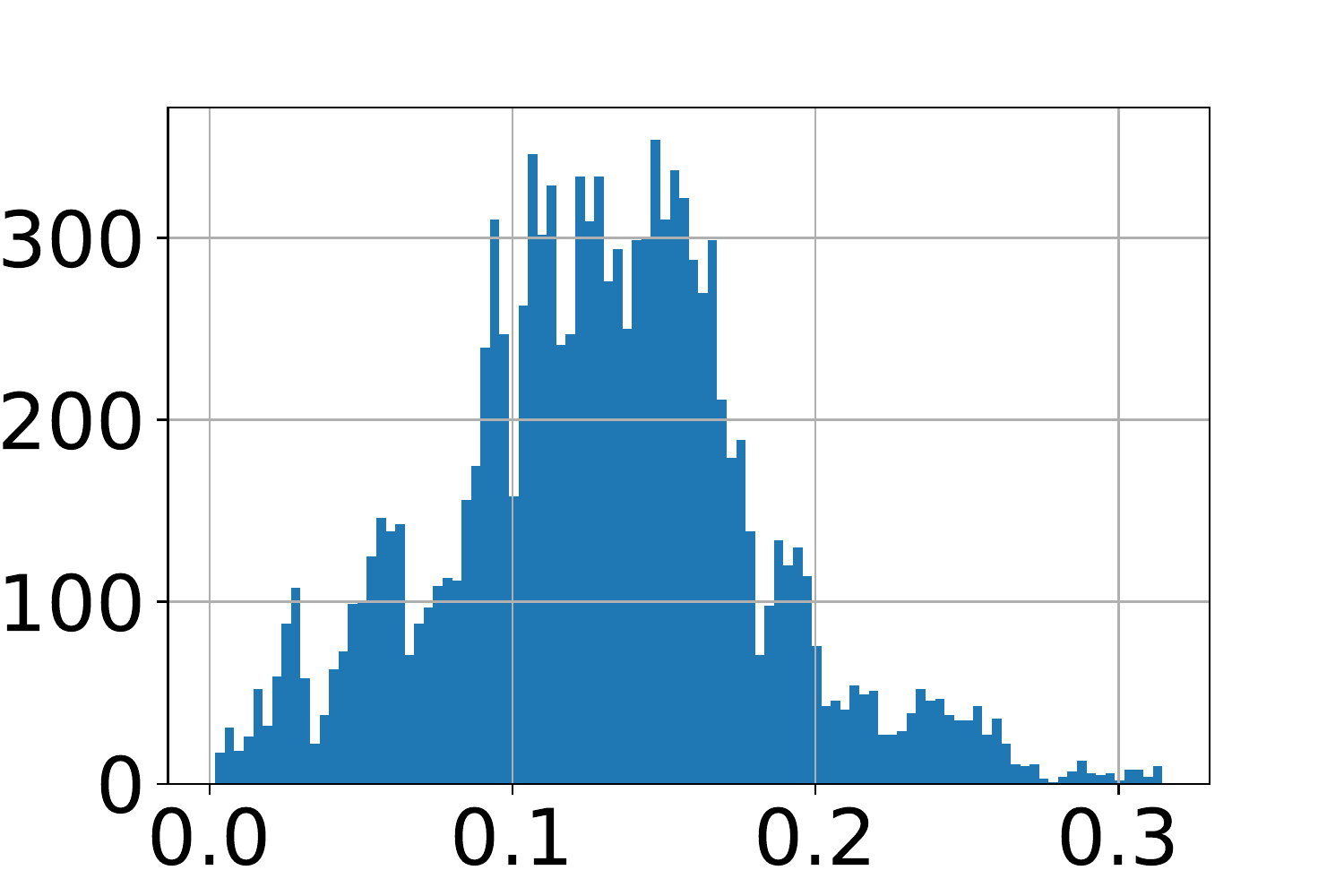}}
  \subfloat[\label{fig:Hist_example_logGaol}logGoal.]
  {\includegraphics[width=0.22\paperwidth]{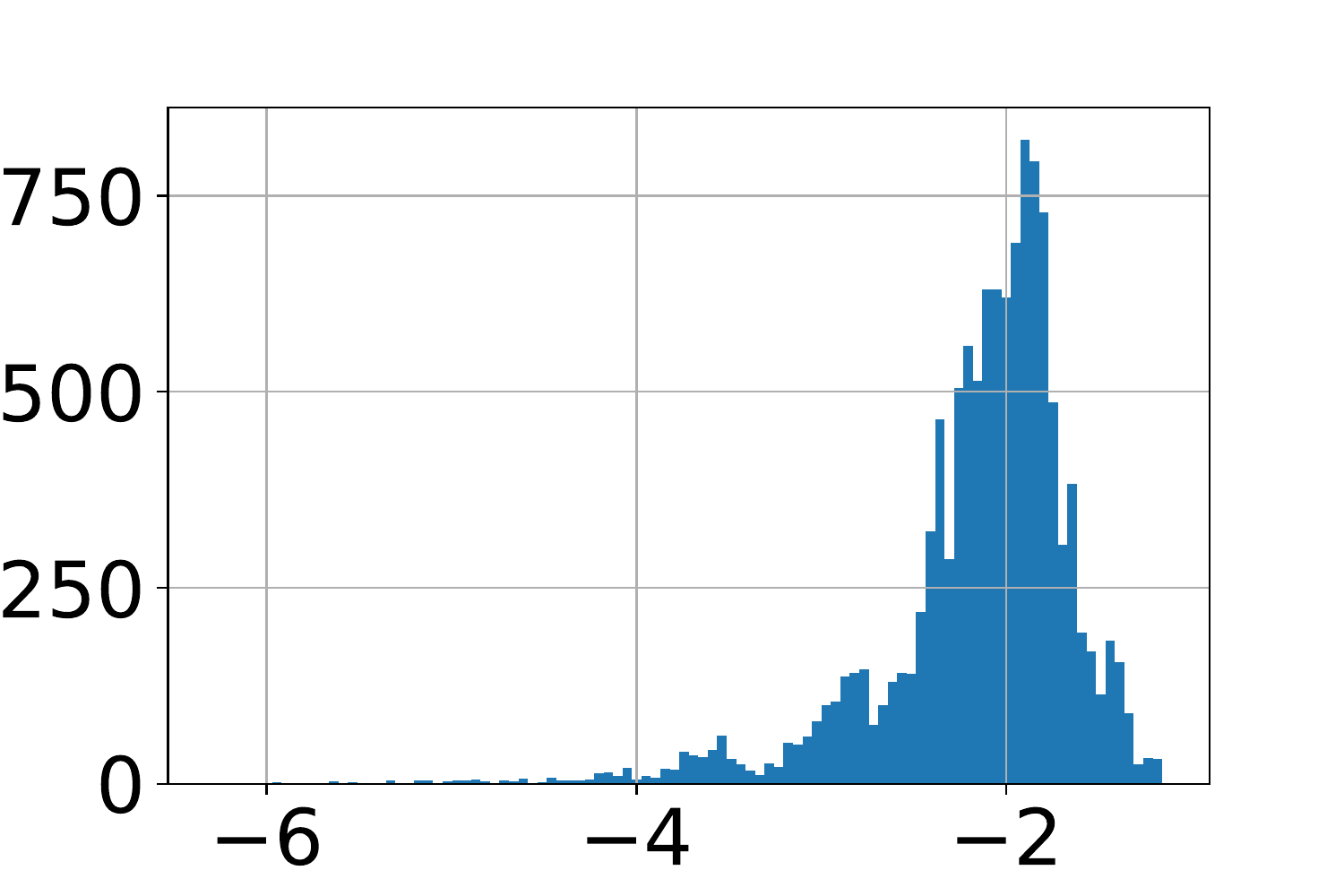}}
  \subfloat[\label{fig:Hist_example_logGaol1000}$\mathrm{log} \left( \mathrm{Gaol} \cdot 1000 \right)$]
  {\includegraphics[width=0.22\paperwidth]{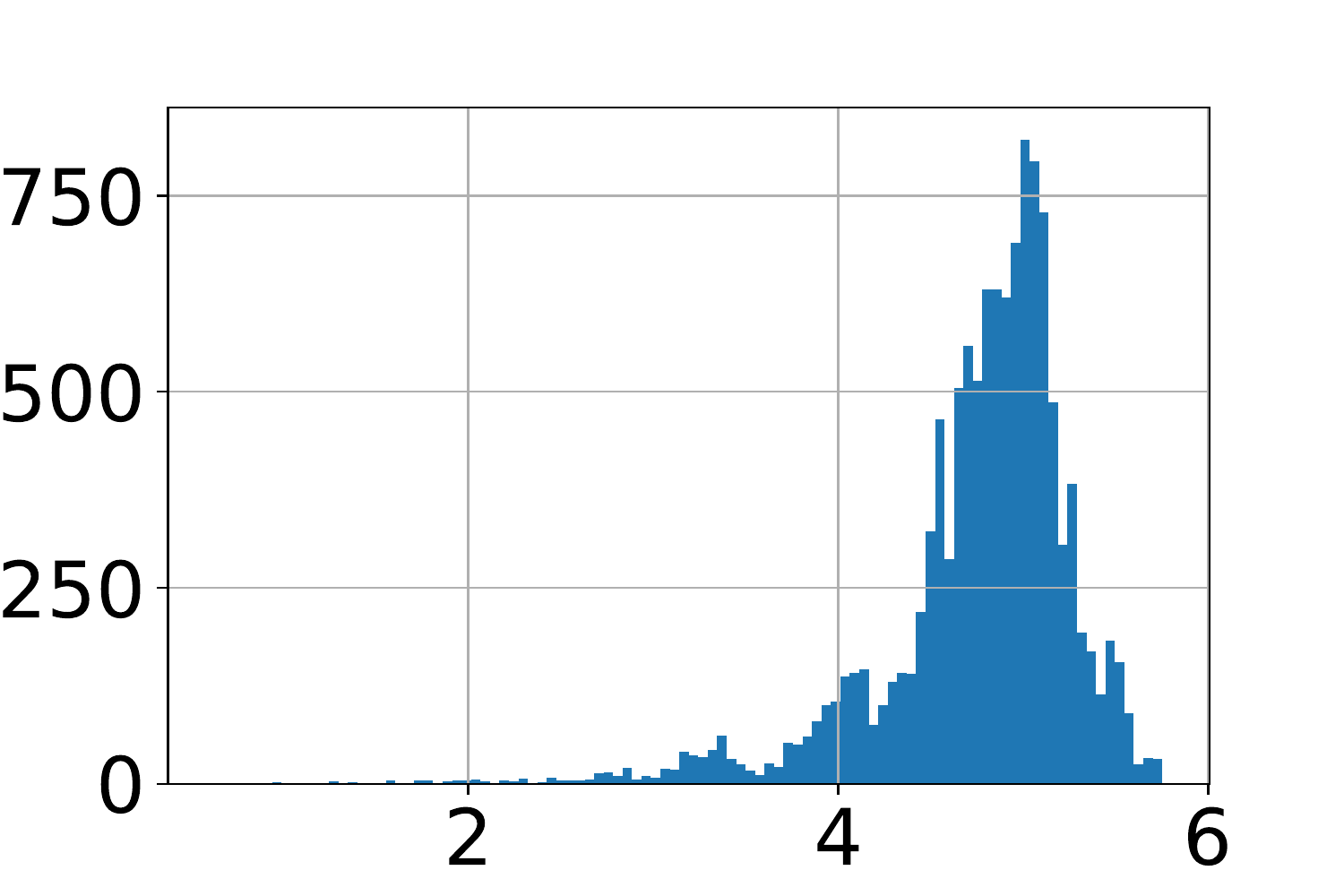}}
  \caption{Illustration of left-skewness and long tails induced by log transform.}
  \label{fig:Hist_example}
\end{figure}

Table~\ref{table:house_variable} also shows the selection results when all variables measured in dollars (e.g., rent, family income, etc.) are transformed by logarithms and the response variable is $\mathrm{log} \left( \mathrm{price} \right)$. The decision to use a log transform only on dollar-measured variables is for both statistical and empirical reasons. Statistically, it is because that the other variables in the data are distributed almost symmetrically without heavy tails. As illustrated with Gaol and Beach in Figure~\ref{fig:Hist_example}, log transforms induce pronounced left skewness. As shown in Figures~\ref{fig:Hist_example_logGaol1000} and \ref{fig:Hist_example_logBeach1000}, left skewness is not resolved by changing variables units before the log transform. Empirically, log transforms may cause interpretation difficulties. For example, we are typically interested in the price response to unit as opposed to a percentage increase in the number of bedrooms or bathrooms. As a result, we do not use log transforms on the other variables.

In log regression, due to variable form changes, lasso selects 35 variables and CV-en selects 54. Some of the lasso and CV-en selections seem odd. For example, lasso drops all Year 5 test scores, Year 3 Spelling and Grammar but selects all the other Year 3 examination scores. CV-en selects all other scores, dropping only Year 5 Reading. These selections seem to suggest that only some primary school examination scores particularly matter in house pricing. By contrast, solar returns a very sparse regression model, with only $11$ variables selected in the log regression and $9$ of them are also selected in the linear regression. Since causal relations should not be affected by variable forms, we select variables chosen by solar simultaneously in Tables~\ref{table:house_variable}: \{bedrooms, baths, Parking, Beach, ChildCare, Gaol, ICSEA, logMortgage, logRent, logFamInc\}.

\begin{table}[h]
  \caption{Regression coefficients: post-selection OLS log model}
  \label{table:coefficients_log}
  \begin{tabular}{@{}l....@{}}
  \toprule

  Variable & \multicolumn{1}{c}{elas net}
           & \multicolumn{1}{c}{lasso}
           & \multicolumn{1}{c}{rec solar}
           & \multicolumn{1}{c}{solar}      \\
  \midrule

  constant    & 8.81\ts   & 8.76\ts   & 7.99\ts   & 7.21\ts  \\

              &  (0.17)   &  (0.15)   &  (0.11)   &  (0.11)  \\

  Bedrooms    & 0.21\ts   & 0.21\ts   & 0.23\ts   & 0.23\ts  \\

              &  (0.00)   &  (0.00)   &  (0.00)   &  (0.00)  \\

  Baths       & 0.09\ts   & 0.10\ts   & 0.09\ts   & 0.09\ts  \\

              &  (0.00)   &  (0.00)   &  (0.00)   &  (0.00)  \\

  Parking     & 0.08\ts   & 0.08\ts   & 0.08\ts   & 0.08\ts  \\

              &  (0.00)   &  (0.00)   &  (0.00)   &  (0.00)  \\

  Airport     & 3.67\ts   & 2.53\ts   & 2.88\ts   &          \\

              &  (0.39)   &  (0.20)   &  (0.25)   &          \\

  Beach       & -1.78\ts  & -2.21\ts  & -1.34\ts  & -2.45\ts \\

              &  (0.31)   &  (0.11)   &  (0.14)   &  (0.14)  \\

  Child care  & -4.39\ts  & -4.49\ts  & -3.63\ts  & -2.45\ts \\

              &  (0.20)   &  (0.16)   &  (0.12)   &  (0.11)  \\

  Gaol        & -0.80\ds  &           & -1.01\ts  & 0.36\ts  \\

              &  (0.32)   &           &  (0.15)   &  (0.14)  \\

  Rubbish     & -0.4      &           & 0.54\ts   &          \\

              &  (0.35)   &           &  (0.21)   &          \\

  log(Mortgage)& 0.16\ts   & 0.16\ts   & 0.24\ts   & 0.26\ts  \\

              &  (0.01)   &  (0.01)   &  (0.01)   &  (0.01)  \\

  log(Rent)   & 0.03\ts   & 0.04\ts   & 0.07\ts   & 0.07\ts  \\

              &  (0.01)   &  (0.01)   &  (0.01)   &  (0.01)  \\

  log(Income) & 0.19\ts   & 0.19\ts   & 0.17\ts   & 0.24\ts  \\

              &  (0.02)   &  (0.01)   &  (0.01)   &  (0.01)  \\

  Age         & 0.01\ts   & 0.01\ts   & 0.01\ts   & 0.01\ts  \\

              &  (0.00)   &  (0.00)   &  (0.00)   &  (0.00)  \\

  ICSEA       & 0.00\ts   & 0.00\ts   & 0.00\ts   & 0.00\ts  \\

              &  (0.00)   &  (0.00)   &  (0.00)   &  (0.00)  \\

  $+$         &           &           &           &          \\

  \midrule

  $p$         &     54    &     36    &     13    &     11    \\

  $R^2$       &      0.77 &      0.76 &      0.74 &      0.73 \\

  $\bar{R}^2$ &      0.77 &      0.76 &      0.74 &      0.73 \\

  $n$         & 11,796    & 11,796    & 11,796    & 11,796    \\

  \bottomrule

  \end{tabular}

\end{table}

\begin{table}[h]

  \caption{Post-selection linear model OLS coefficients for variables selected by rectified solar.}

  \label{table:coefficients_linear}

  \begin{tabular}{@{}l....@{}}

  \toprule

  Variable & \multicolumn{1}{c}{elas net}

           & \multicolumn{1}{c}{lasso}

           & \multicolumn{1}{c}{rec solar}

           & \multicolumn{1}{c}{solar} \\
  \midrule

  constant   &  -886234.40\ts &  -827387.49\ts & -1445430.40\ts & -2486422.15\ts \\

             &  (186680.91)   &  (174136.59)   &  (112390.24)   &   (98422.98)   \\

  Bedrooms   &   165639.00\ts &   166225.82\ts &   183893.59\ts &   169510.52\ts \\

             &    (6433.57)   &    (6404.82)   &    (6015.93)   &    (6015.37)   \\

  Baths      &   210101.84\ts &   210600.58\ts &   203674.28\ts &   209626.52\ts \\

             &    (8115.07)   &    (8048.92)   &    (8147.93)   &    (8297.24)   \\

  Parking    &    97790.57\ts &    96883.13\ts &   104050.40\ts &    97623.23\ts \\

             &    (6689.35)   &    (6688.13)   &    (6861.16)   &    (6985.67)   \\

  Airport    &  2865246.35\ts &  3122719.74\ts &  1849108.51\ts &                \\

             &  (735335.86)   &  (625604.38)   &  (454344.97)   &                \\

  Beach      & -5029681.74\ts & -4051671.57\ts & -1509612.24\ts &  -796281.77\ts \\

             &  (600061.28)   &  (262369.00)   &  (260370.04)   &  (153431.50)   \\

  Child care & -4802095.18\ts & -4163486.91\ts & -3961629.37\ts &                \\

             &  (393577.54)   &  (316107.89)   &  (220752.57)   &                \\

  Gaol       &  1614215.27\ds &                & -1137143.99\ts & -1909369.80\ts \\

             &  (646392.84)   &                &  (267597.21)   &  (107204.64)   \\

  Rubbish    &   -45084.93    &   780180.22    &  3136997.85\ts &                \\

             &  (672532.42)   &   594084.58)   &  (372355.34)   &                \\

  Mortgage   &      133.67\ts &      134.18\ts &      174.55\ts &      185.99\ts \\

             &       (7.98)   &       (7.95)   &       (7.82)   &       (7.96)   \\

  Rent       &      264.35\ts &      265.80\ts &      312.85\ts &      370.76\ts \\

             &      (36.80)   &      (35.27)   &      (34.78)   &      (35.42)   \\

  Income     &       59.04\ts &       69.22\ts &       -8.39    &       66.57\ts \\

             &      (19.23)   &      (14.97)   &      (12.18)   &      (11.48)   \\

  Age        &     3673.46\ts &     4106.29\ts &                &                \\

             &    (1031.69)   &      (958.98)  &                &                \\

  ICSEA      &      838.54\ts &      862.92\ts &      960.68\ts &     1756.92\ts \\

             &     (172.42)   &     (163.42)   &     (104.07)   &      (95.67)   \\

  $+$        &                &                &                &                \\

  \midrule

  $p$         &       57      &        44      &       12       &        9       \\

  $R^2$       &        0.548  &         0.548  &        0.514   &        0.494   \\

  $\bar{R}^2$ &        0.546  &         0.546  &        0.514   &        0.493   \\

  $n$         &   11,974      &     11,974     &   11,974       &   11,974       \\

  \bottomrule

  \end{tabular}

\end{table}

Lastly, solar variable selection outperforms lasso-type estimators in terms of the balance between sparsity and prediction accuracy, as shown in Tables~\ref{table:coefficients_log} and \ref{table:coefficients_linear}. Table~\ref{table:coefficients_linear} details the post-selection OLS results on CV-en, lasso and solar selection from the log models, showing solar selects only 9 variables compared with lasso (44) and CV-en (57). Surprisingly, pruning 35 to 48 variables from the \ref{table:house_variable} list only reduces the $R^2$ by 5\%. This confirms that, from the perspective of prediction, solar successfully identifies the most important variables in the database. A very similar result is also found in Table~\ref{table:coefficients_log} where solar only selects 11 out of 57 variables, which explains 73\% of the variation of log(price). The extra variables selected by lasso or CV-en only improve the $R^2$ by around 2\%. It is known that more covariates, redundant or not, always increase $R^2$ in finite. In this datset, 25 more variables (from solar to lasso) only increases $R^2$ by 3\%. This suggests that either the 3\% gain is pure overfitting; or they are conditionally correlated to $\log \left( \mathrm{Price} \right)$ in a very weak level. As those with direct causal relations from/to $\log \left( \mathrm{Price} \right)$, the Markov Blanket variables of $\log \left( \mathrm{Price} \right)$ are always strongly correlated to $\log \left( \mathrm{Price} \right)$, holding all other variables constant. A weak conditional correlation suggest that these variables may be the remote ancesters/descendants of $\log \left( \mathrm{Price} \right)$. This does not suggest that they have no causal relations to $\log \left( \mathrm{Price} \right)$; it just implies that they are not in the MB of $\log \left( \mathrm{Price} \right)$ and do not have a direct causal effect. 

While the log and linear models should represent the same causal structure, the linear model performs relatively poorly. Also, with log transform, the dollar measured variables are less skewed and possibly with a lighter tail. Hence, we focus on the log regression. As explained previously, the MB includes all the variables that are conditionally correlated to price in the population, implying that the MB variables should be able to explain all non-noise variation in price. Since we do not know the population variance of noise, we cannot know with absolute certainty the magnitude of noise variation. However, with $R^2=73$\%, we are confident that the majority of price variation is linear and explained by the MB variables. The remaining 27\% may be due to noise, functional form error (e.g., to capture nonlinear patterns, we should use a polynomial equation, a trigonometric equations or a nerual network instead of a first-order linear equation), or spatial clustering in the geographical data. While we cannot rule out nonlinearity, the severity of those problems appears to be under control. The high explanatory power of the variables selected by solar under the log transform is reassuring on the reliability of MB selection. In Appendix~3, we try using other machine learning methods to capture the nonlinearity among $\log \left( \mathrm{price} \right)$ and the selected $11$ variables. It turns out that, with cross validation controlling the overfitting, the selected variable can easily explain around 90\% of the variation of $\log \left( \mathrm{price} \right)$, confirming that there does exist a nonlinear pattern between the selected variables and $\log \left( \mathrm{price} \right)$, which accounts for another 17\% of $R^2$ and is not the major pattern in our dataset.

\subsection{Grouping effects in variable selection}

Before moving on to graph learning, we need to check whether grouping effects cause solar to mistakenly exclude variables from the price MB. As noted above, the accuracy and robustness of variable selection may be reduced when grouping effects are embedded in the data, especially when the dependence and causation structures are complicated. As shown in the supplementary correlation table, the distances of houses to different locations are highly spatially correlated with one another. To investigate whether solar variable selection is affected by such multicollinearity, Table~\ref{table:corr_gaol} focuses on the group of variables highly correlated to gaol, including airport, rubbish and childcare, all of which have pairwise unconditional correlations above 0.5.

\begin{table}[H]
 \centering
 \caption{Unconditional correlations to Gaol (absolute value $> 0.5$) \label{table:corr_gaol}}
 \begin{tabular}{lcccc.}
  \toprule
    & ChildCare & Airport & Rubbish & Beach \\
  \midrule
  $\mathrm{corr} \left( \; \cdot \; , \mathrm{Gaol} \right)$ & 0.756 & 0.715 & 0.671 & 0.528 \\
  \bottomrule
 \end{tabular}
\end{table}

Based on the results in Table~\ref{table:corr_gaol}, we standardize all variables and estimate the regression
\begin{equation}
 \mathrm{Gaol} = \gamma_0 + \gamma_1 \cdot \mathrm{Airport} + \gamma_2 \cdot \mathrm{ChildCare} + \gamma_3 \cdot \mathrm{Rubbish} + \gamma_4 \cdot \mathrm{Beach} + e.
 \label{OLS:Gaol_others}
\end{equation}
The estimation results from (\ref{OLS:Gaol_others}) are in Table~\ref{table:OLS_gaol}.

\begin{table}[H]
 \centering
 \caption{OLS results from (\ref{OLS:Gaol_others}) \label{table:OLS_gaol}}
\begin{tabular}{r....}
  \toprule
            & \multicolumn{1}{c}{coefficient}
            & \multicolumn{1}{c}{$SE$}
            & \multicolumn{1}{c}{$t $}
            & \multicolumn{1}{c}{$P>|t|$} \\
  \midrule
  constant  & 0      & 0.003   & 0       & 1.000 \\
  Airport   & 0.4488 & 0.011   & 41.063  & 0.000 \\
  ChildCare & 0.3276 & 0.006   & 56.908  & 0.000 \\
  Rubbish   & 0.0373 & 0.010   & 3.849   & 0.000 \\
  Beach     & 0.5522 & 0.003   & 174.257 & 0.000 \\
  \bottomrule
 \end{tabular}
 \begin{tabular}{r.r.}
  $n$          & 11,974 & $F$                & 23,110 \\
  $R^2$        & 0.885  & $P(F)$             & 0     \\
  $\bar{R}^2$  & 0.885  & degrees of freedom & 4     \\
  \bottomrule
\end{tabular}
\end{table}

Table~\ref{table:OLS_gaol} shows that almost 90\% of the variation in Gaol can be explained by \{ChildCare, Airport, Rubbish, Beach\} and $\sum_{\forall i \neq 0} \left\vert \gamma_i \right\vert = 1.35$ in (\ref{OLS:Gaol_others}), indicating severe multicollinearity between Gaol and the other 4 variables. The empirical reason why \{Gaol, ChildCare, Airport, Rubbish, Beach\} are highly correlated is easy to see. The house market data cover a roughly 10km square area in eastern Sydney. The gaol (Long Bay correctional complex), several childcare centers (e.g., Blue Gum Cottage Child Care, Alouette Child Care, etc.), the airport (Kingsford-Smith Airport) and waste treatment facilities (e.g., Banksmeadow Transfer Terminal, Malabar Wastewater Treatment Plant, Cronulla Wastewater Treatment Plant) are all located in the southeast corner of the 10km square area, explaining the collinearity among the variables.

The multicollinearity very likely breaches the irrepresentable condition (IRC) and indicates the presence of grouping effects, casting doubt on the variable selection process. The implication is that, even though we know that at least one of the variables in \{ChildCare, Airport, Rubbish, Beach, Gaol\} is in the MB of price, it may be difficult to pinpointing precisely which one statistically. To avoid being misled by any grouping effect, we consider enlarge the subset \{Gaol, ChildCare, Beach\} into the \textbf{gaol group} \{ChildCare, Airport, Rubbish, Beach, Gaol\} in the variable selection results. We refer to the union of the solar variables and \{ChildCare, Airport, Rubbish, Beach, Gaol\} as the \textbf{rectified solar selection}. Thus, for completeness, we compare the OLS results in both linear and log forms with the selection results from lasso, CV-en, solar ((\ref{OLS:solar}) and (\ref{OLS:solar_log})) and rectified solar selection ((\ref{OLS:postsolar}) and (\ref{OLS:postsolar_log})).
\begin{align}
  \mathrm{Price} = \beta_0 & + \beta_1 \cdot \mathrm{Mortgage} + \beta_2 \cdot \mathrm{Rent} +
  \beta_3 \cdot \mathrm{FamInc} + \beta_4 \cdot \mathrm{Bedrooms} \label{OLS:solar} \\
  & + \beta_5 \cdot \mathrm{Baths} + \beta_6 \cdot \mathrm{Parking} + \beta_7 \cdot \mathrm{Beach} + \beta_8 \cdot \mathrm{Gaol} + \beta_9 \cdot \mathrm{ICSEA} + u; \notag \\
  \mathrm{Price} = \beta_0 & + \beta_1 \cdot \mathrm{Mortgage} + \beta_2 \cdot \mathrm{Rent}
  + \beta_3 \cdot \mathrm{FamInc} + \beta_4 \cdot \mathrm{Bedrooms}  \label{OLS:postsolar} \\
  & + \beta_5 \cdot \mathrm{Baths} + \beta_6 \cdot \mathrm{Parking}  + \beta_7 \cdot \mathrm{Beach} + \beta_8 \cdot \mathrm{Airport} + \beta_9 \cdot \mathrm{ChildCare} \notag   \\
  & + \beta_{10} \cdot \mathrm{Rubbish} + \beta_{11} \cdot \mathrm{ICSEA} + u; \notag
\end{align}
\begin{align}
  \mathrm{logPrice} = \beta_0 & + \beta_1 \cdot \mathrm{logMortgage} +
  \beta_2 \cdot \mathrm{logRent} + \beta_3 \cdot \mathrm{logFamInc} + \beta_4 \cdot \mathrm{Bedrooms} \label{OLS:solar_log} \\
  & + \beta_5 \cdot \mathrm{Baths} + \beta_6 \cdot \mathrm{Parking} + \beta_7 \cdot \mathrm{Beach} + \beta_8 \cdot \mathrm{Gaol} + \beta_9 \cdot \mathrm{ICSEA} + u; \notag \\
  \mathrm{logPrice} = \beta_0 & + \beta_1 \cdot \mathrm{logMortgage}
  + \beta_2 \cdot \mathrm{logRent}
  + \beta_3 \cdot \mathrm{logFamInc} + \beta_4 \cdot \mathrm{Bedrooms}  \label{OLS:postsolar_log} \\
  & + \beta_5 \cdot \mathrm{Baths} + \beta_6 \cdot \mathrm{Parking}  + \beta_7 \cdot \mathrm{Beach} + \beta_8 \cdot \mathrm{Airport} + \beta_9 \cdot \mathrm{ChildCare} \notag   \\
  & + \beta_{10} \cdot \mathrm{Rubbish} + \beta_{11} \cdot \mathrm{ICSEA} + u. \notag
\end{align}

The comparisons are also summarized in Tables~\ref{table:coefficients_log} and \ref{table:coefficients_linear}. The most interesting thing is the $R^2$ from rectified solar. The difference between the solar and CV-en $R^2$ values tells us that the 48 variables dropped by solar explain a mere 5\% of price variation while the difference between the solar and rectified solar $R^2$ shows that the gaol group dropped by solar explains 2\% of price variation. A very similar result can be found in the $R^2$ comparison of log models. Thus, among all the 48 dropped variables, \{Airport, Rubbish\} seem to be the most important. This is additional evidence to justify previous doubts about the grouping effect.

\section{Score-based graph learning based on solar variable selection}

In last section, we select MB members of house prices using rectified solar. Ceteris paribus, each variable selected by rectified solar is highly likely to be conditionally correlated to price in the population, implying that they are highly likely to be the MB of price. However, it is possible that these variables have different roles: some may serve as the parents of price while others may serve as children or spouses. In order to accomplish endogeneity detection and instrument variable selection, we need to estimate the role of each MB member and all the complete pattern of causations in the MB. For this step, we implement the score-based graph learning method.

\subsection{Temporal ordering of MB variables and Markov equivalence}

\begin{figure}[H]
  \centering
  \subfloat[\label{fig:Markov_Equi1}population graph]
  {\includegraphics[width=0.2\paperwidth]{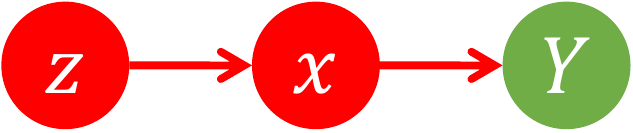}}
  \hfil
  \subfloat[\label{fig:Markov_Equi2}skeleton of the population graph]
  {\includegraphics[width=0.2\paperwidth]{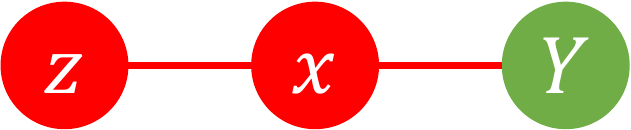}}
  \hfil
  \subfloat[\label{fig:Markov_Equi3}Markov equivalence 1]
  {\includegraphics[width=0.2\paperwidth]{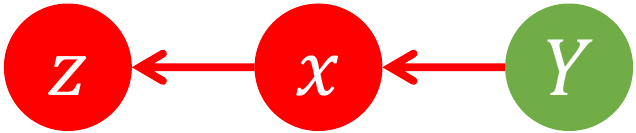}}

  \subfloat[\label{fig:Markov_Equi4}Markov equivalence 2]
  {\includegraphics[width=0.25\paperwidth]{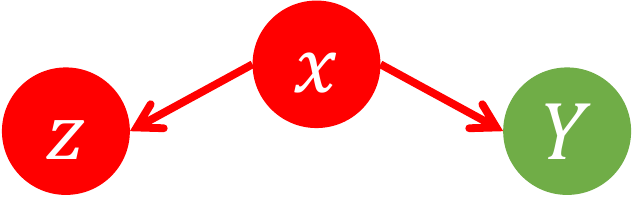}}
  \hfil
  \subfloat[\label{fig:Markov_Equi5}non-Markov-equivalence]
  {\includegraphics[width=0.25\paperwidth]{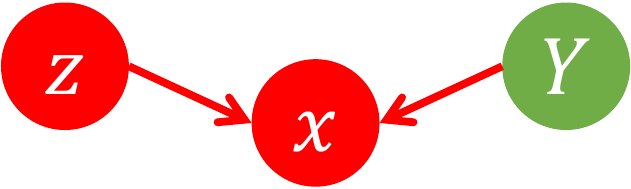}}
  \caption{Illustration of Markov equivalence.}
  \label{fig:Markov_Equi}
\end{figure}

A common problem in graph learning and causal inference is the Markov equivalence. In a nutshell, Markov equivalence says that, without an exact \textbf{time stamp} (i.e., when a variable is generated or, equivalentlym, when the value of a variable is determined), we cannot learn the exact population graph from the data. Instead, we can only learn the skeleton of the population graph (i.e., a graph without arrows or an undirected graph). Figure~\ref{fig:Markov_Equi} illustrates these concepts in a simple example.

Figure~\ref{fig:Markov_Equi1} shows the population graph to be $\mathbf{z} \rightarrow \mathbf{x} \rightarrow \mathbf{y}$. However, without knowing the time stamp of each variable (which variable is born first), it is impossible to find the correct parent-child relations. We can only find a skeleton graph (Figure~\ref{fig:Markov_Equi2}), where we know that $\mathrm{corr} \left( \mathbf{z}, \mathbf{x}\right) \neq 0$, $\mathrm{corr} \left( \mathbf{z}, \mathbf{y} \right) \neq 0$, $\mathrm{corr} \left( \mathbf{y}, \mathbf{x}\right) \neq 0$ and $\mathrm{corr} \left( \mathbf{y} \vert \mathbf{x}, \ \mathbf{z} \vert \mathbf{x} \right) = 0$. Any graphs that fit these conditions are included in the Markov equivalence class. Thus, Figure~\ref{fig:Markov_Equi1} (population graph), Figure~\ref{fig:Markov_Equi3} and Figure~\ref{fig:Markov_Equi4} (the confounding or fork structure) are included in the Markov equivalence class. Figure~\ref{fig:Markov_Equi5} (the collider structure) is not included since the correlation between $\mathbf{z}$ and $\mathbf{y}$ is zero unless $\mathbf{x}$ is conditioned on. Put another way, without specific time stamps in this example, graph learning can only identify whether or not the population graph follows a collider structure, which is not particularly useful. However, with the corresponding time stamps and a large number of variables, we can eliminate Markov equivalence and narrow the list of candidates in the population graph. For example, if we know that $\mathbf{z}$, $\mathbf{x}$ and $\mathbf{y}$ were generated at respectively year 1990, 1991 and 1992, we can rule out Figure~\ref{fig:Markov_Equi3} and \ref{fig:Markov_Equi4} from the Markov Equivalence, implying that Figure~\ref{fig:Markov_Equi1} is the correct graph.

\begin{figure}[H]
  \centering
  \includegraphics[width=0.6\paperwidth]{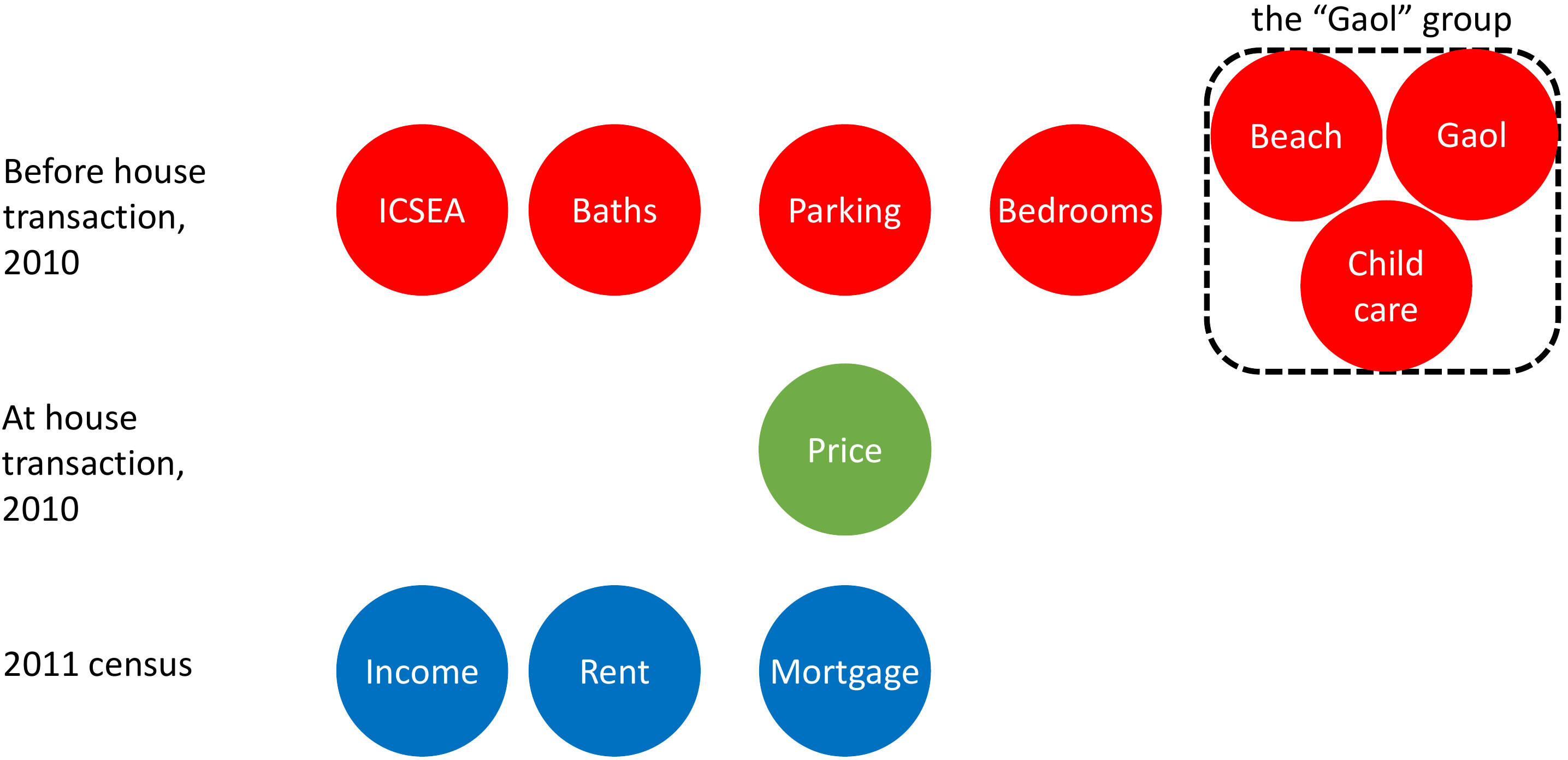}
  \caption{Chronology of variables in the MB of price.}
  \label{fig:BN_pre}
\end{figure}

\begin{figure}[h]
  \centering
  \includegraphics[width=0.6\paperwidth]{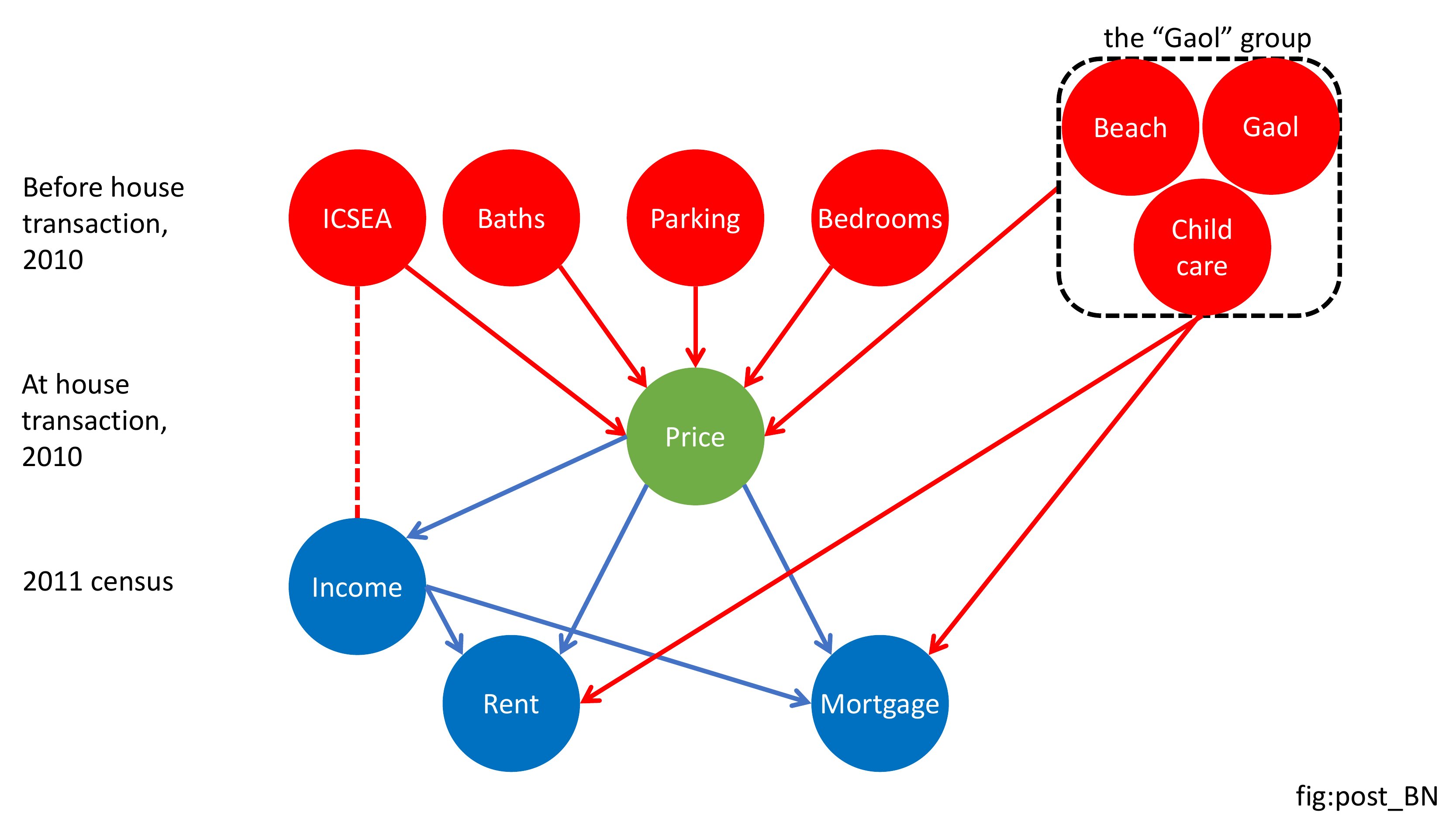}
  \caption{The estimated graph based on solar variable selection.}
  \label{fig:BN_post}
\end{figure}

Based on the temporal order of variable generation in the house data, the selected variables can be ordered vertically. In Figure~\ref{fig:BN_pre}, the red nodes (house features, distances and ICSEA) are determined before the house transaction in 2010; the green nodes are generated at the time of the 2010 house transaction; and the blue nodes (demographic variables) are generated at the 2011 census, after 2010 house transactions.\footnote{Multiple green nodes are included in our database, including the house transaction method (e.g., auction, private sale, and so on), the listing history and 30 other variables. We focus on price in this graph estimation.} Due to the probable IRC violation discussed above, \{Gaol, ChildCare, Beach\} are grouped manually and represent the the `Gaol group' \{Gaol, ChildCare, Beach, Airport, Rubbish\}. The temporal order helps to identify the role of each variable in the MB: variables generated in 2011 cannot cause any change in those generated in or before 2010, implying that the red nodes cannot be the descendants of green and blue nodes; likewise, the green nodes cannot be the children of blue nodes as well. Since all selected variables are in the MB of price, the red nodes have to be the parents of price while rent and mortgage are the children of price. Since parents cause their children, who further cause the grandchildren, the time stamps and MB variable yield the estimated graph and the causation realtions as Figure~\ref{fig:BN_post}.

The role of income in Figure~\ref{fig:BN_post} is worthy of discussion. 2009 ICSEA is computed partially based on household income at 2009 while the income variable is household income in 2011. Since there exists strong temporal correlation in household income and since we do not have a high-dimensional database on household income, we cannot determine whether the correlation between ICSEA and income is purely autocorrelation or contains some kind of causation. Assuming there exists causation, we cannot identify it statistically without detailed earnings data linked to house transactions. Detailed earnings data are beyond our scope. Moreover, linking detailed earnings data to the house addresses in the transactions data would likely run afoul of data privacy regulations. Hence, we connect ICSEA and income with an undirected dash, meaning graphically that there may be some kind of relation that we cannot identify in detail. Since solar variable selection confirms that income and the gaol group belong to both the MB of rent and MB of mortgage (details can be found in the supplementary files), we connect them to rent and mortgage directly, which completes the  graph learning as Figure~\ref{fig:BN_post}.

\subsection{Backdoor effect estimation}

\begin{figure}[H]
  \centering
  \subfloat[\label{fig:Backdoor_bath1} subgraph without BE]
  {\includegraphics[width=0.18\paperwidth]{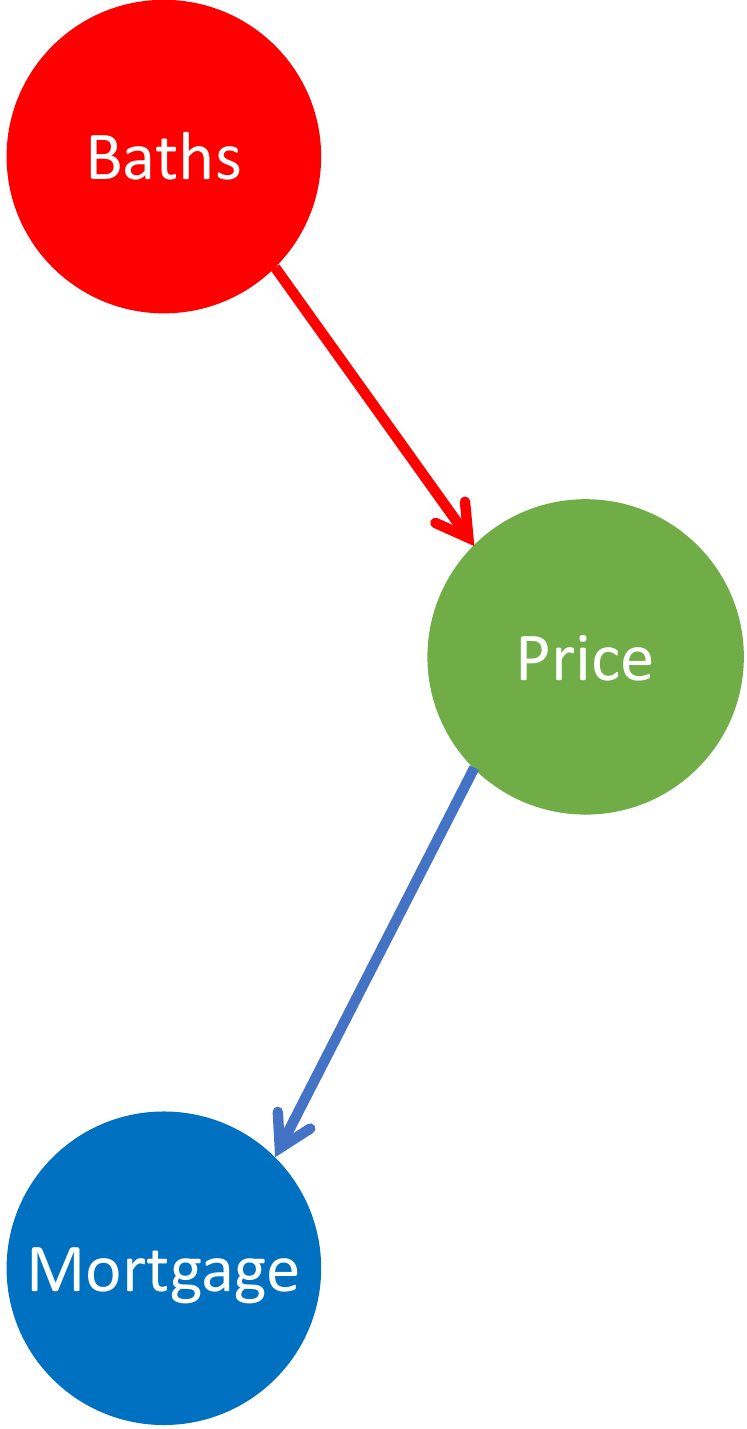}}
  \hfil
  \subfloat[\label{fig:Backdoor_bath2} subgraph with BE]
  {\includegraphics[width=0.18\paperwidth]{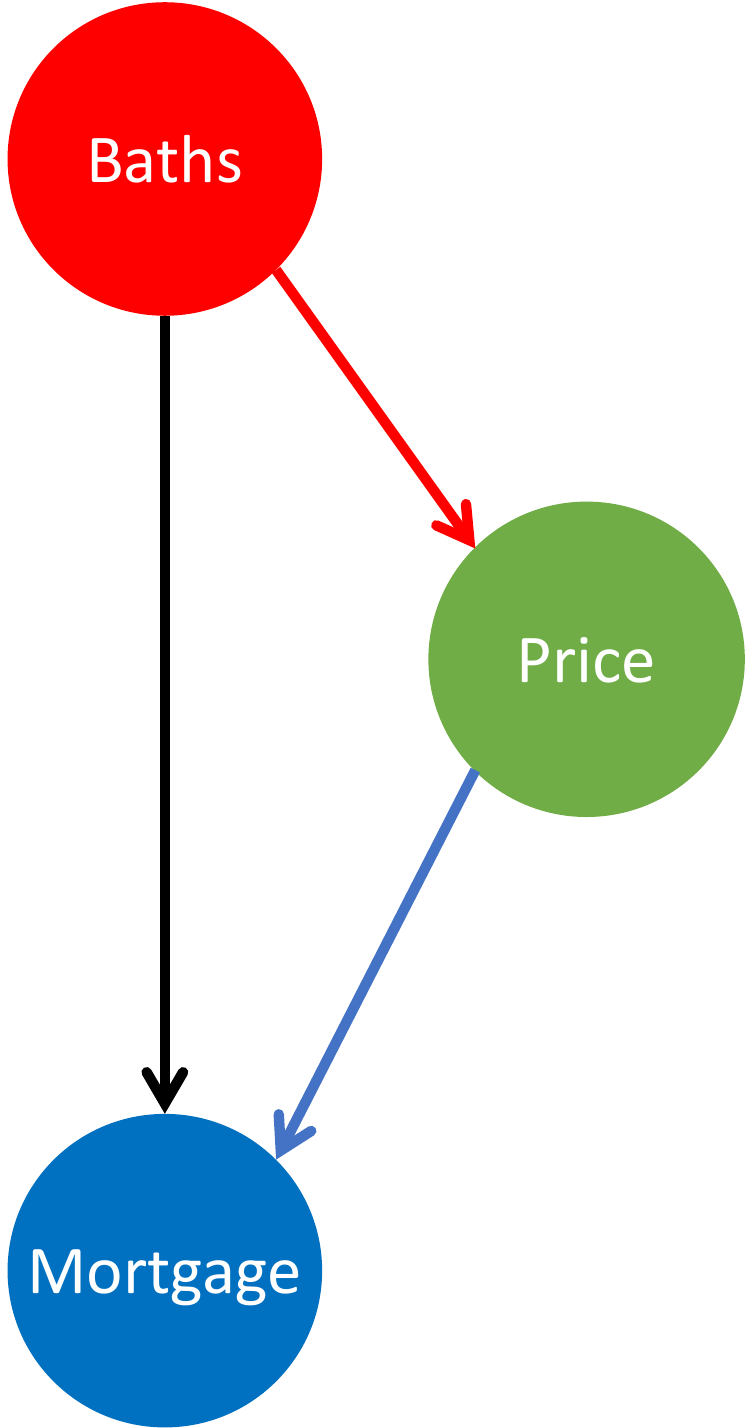}}
  \hfil
  \subfloat[\label{fig:Backdoor_bath3} alternative subgraph with BE]
  {\includegraphics[width=0.18\paperwidth]{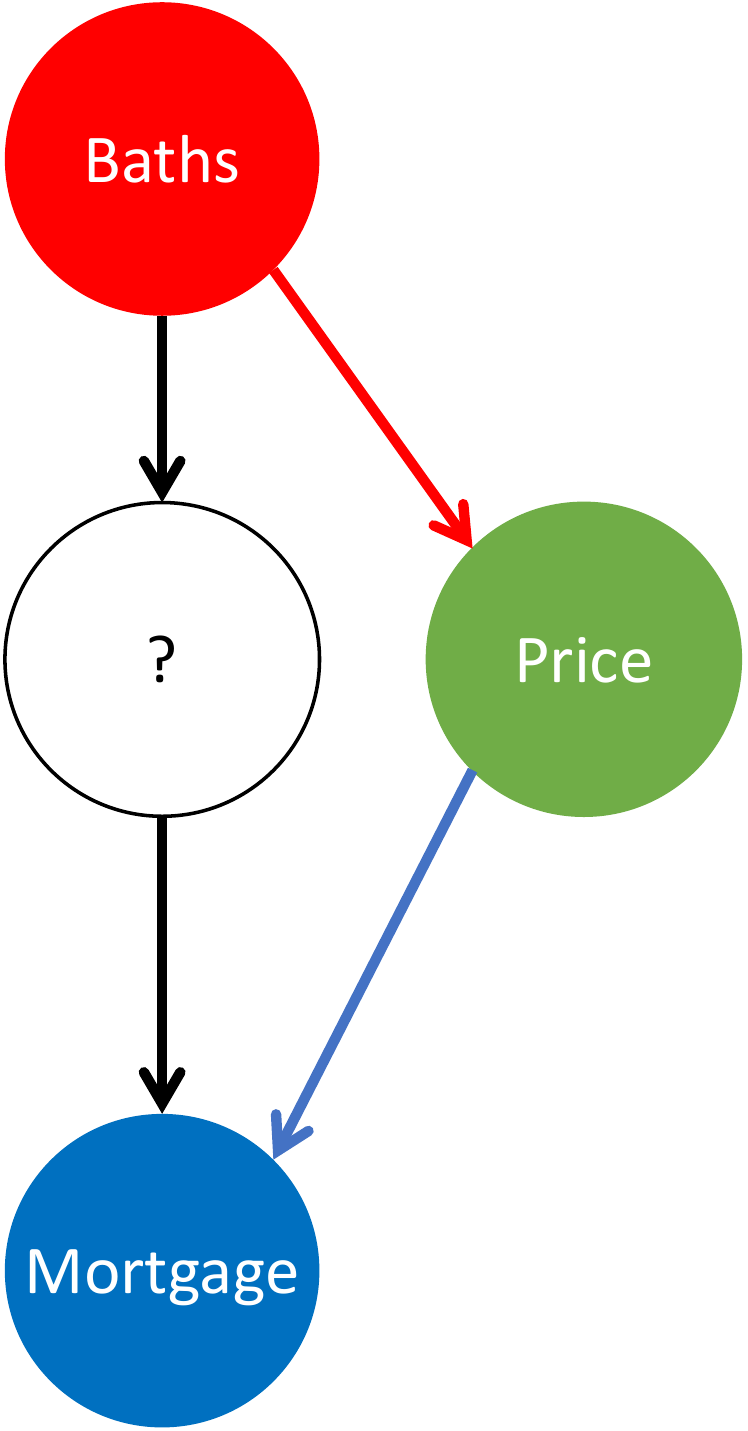}}
  \caption{Illustration of BE estimation between baths, price and mortgage.}
  \label{fig:Backdoor_bath}
\end{figure}

Figure~\ref{fig:BN_post} shows that a parent of price can indirectly cause a change in rent or mortgage through price (e.g., $\mathrm{Bath} \rightarrow \mathrm{Price} \rightarrow \mathrm{Mortgage}$), confirming the existence of the frontdoor effect(FE). The last step of graph learning is to determine whether the parents of price can cause a change in rent or mortgage without going through price. Such a causal effect is a backdoor effect (BE) and illustrated as black arrows in Figure~\ref{fig:Backdoor_bath}. BE can be constructed either as Figure~\ref{fig:Backdoor_bath2} or Figure~\ref{fig:Backdoor_bath3}. The difference between Figures~\ref{fig:Backdoor_bath2} and \ref{fig:Backdoor_bath3} is that, in Figure~\ref{fig:Backdoor_bath2}, Baths can cause a change in Mortgage even after controlling for any possible variables. BEs of such kind cannot be cut off even though you control all the other variables in the world. By contrast, after controlling the variable(s) `?' and Price in Figure~\ref{fig:Backdoor_bath3}, Baths can no longer cause any change in Mortgage, meaning that BEs of such kind are controllable. We can estimate which causal structure fits the data better as follows. First, we find out which of Figures~\ref{fig:Backdoor_bath1} and \ref{fig:Backdoor_bath2} fits data better without controlling for any other variables. If Figure~\ref{fig:Backdoor_bath1} is chosen, it implies that there is no BE. If Figure~\ref{fig:Backdoor_bath2} is chosen, there exists a BE and we need to combinatorially determine whether there exists a variable(s) `?' in the BE from Baths to Mortgage.

We use the score-based learning method to find the optimal causal structure. Specifically, we estimate the AIC, BIC and BGE scores of Figure~\ref{fig:Backdoor_bath1} (column `no BE' in Table~\ref{Table:Backdoor}) and Figure~\ref{fig:Backdoor_bath2} (column `BE' in Table~\ref{Table:Backdoor}) on the house pricing data and choose the one with the lower score. If bath only causes mortgage via price, the conditional correlation between baths and mortgage will be very close to zero after holding price constant. Hence, due to the overfitting of the causal structure in Figure~\ref{fig:Backdoor_bath2}, its AIC, BIC or BGE scores will be higher than for Figure~\ref{fig:Backdoor_bath1}, implying that the no-BE graph fits the data better. By replacing baths with parking or bedrooms in Figure~\ref{fig:Backdoor_bath}, we can instead check whether a BE exists between mortgage and other parents of price. In a similar vein, by replacing mortgage with rent, we can also check whether a BE exists between rent and any parent of price. Table~\ref{Table:Backdoor} shows the results from BE estimation.

\begin{table}[H]
  \caption{Estimation of the BE between the parents and children of price\\ (optimal AIC, BIC, and BGE scores in red)}
  \label{Table:Backdoor}
  \begin{tabular}{cll>{\color{red}}ll>{\color{red}}l}
    \toprule
    & & \multicolumn{2}{c}{logRent}
      & \multicolumn{2}{c}{logMortgage} \\ \cline{3-6}
    \\ [-8pt]
    & & \multicolumn{1}{c}{BE}
      & \multicolumn{1}{c}{No BE}
      & \multicolumn{1}{c}{BE}
      & \multicolumn{1}{c}{No BE} \\
    \midrule
             & AIC & $-$23726.55 & $-$23731.28 & $-$19437.42 & $-$19482.37 \\
    Baths    & BIC & $-$23759.74 & $-$23760.78 & $-$19470.61 & $-$19511.87 \\
             & BGE & $-$23758.70 & $-$23760.01 & $-$19470.54 & $-$19512.35 \\ \cline{1-6}
             & AIC & $-$27197.09 & $-$27500.08 & $-$23060.65 & $-$23251.17 \\
    Bedrooms & BIC & $-$27230.27 & $-$27529.58 & $-$23093.84 & $-$23280.67 \\
             & BGE & $-$27230.75 & $-$27529.87 & $-$23095.31 & $-$23282.22 \\ \cline{1-6}
             & AIC & $-$25002.25 & $-$25064.54 & $-$20762.16 & $-$20815.63 \\
    Parking  & BIC & $-$25035.44 & $-$25094.04 & $-$20795.35 & $-$20845.13 \\
             & BGE & $-$25034.62 & $-$25093.35 & $-$20795.46 & $-$20845.70 \\
  \bottomrule
  \end{tabular}
\end{table}

We use logRent and logMortgage for the BE estimation in Table~\ref{Table:Backdoor} because, although the score-based learning method works well on many subgaussian variables for small $p/n$, Rent and mortgage are typically right-skewed distributions. The log transform can significantly reduce right skewness and improve the accuracy of the AIC, BIC, and BGE scores. Table~\ref{Table:Backdoor} clearly shows that the graph without BEs consistently have lower scores in terms of AIC, BIC and BGE, which confirms the validity of no BE. As a result, Figure~\ref{fig:BN_post} is the final estimated graph as the MB of price; we don't need to add in any BE.

It is worth noting that we ignore ICSEA, income and the gaol group in Table~\ref{Table:Backdoor} for statistical reasons. Due to IRC violation in the gaol group, it is difficult to estimate accurately the BE of variables in the group to the children of price.\footnote{Statistically, the minimal eigenvalue of the covariance matrix may be very close to $0$, resulting in unreliable AIC, BIC, or BGE scores.} ICSEA is synthesized from a number of variables such as household income, family wealth and other factors representing the socio-economic status of the household. As a result, there may exists a complicated causal relation among each part of ICSEA and price, which we cannot investigate due to the unknown form of the synthesis.

\subsection{Graph interpretations and remarks on AreaSize}

Figure~\ref{fig:BN_post} shows that the graph estimation and MB selection offers interpretations consistent with economic intuition about the dynamics of the house market. First, houses with more desirable features and better locations are purchased by higher-income households at higher prices. Second, higher house prices causes higher mortgage repayments and higher rental payments. Third, higher-rent houses are leased to higher-income households (households with high ICSEA scores). From a demographic perspective, after higher-income home owners or tenants move into the newly purchased or leased houses, the average income and family income in the local SA1 also increases, reflected on the graph as price causing income.

\begin{figure}[H]
  \centering
  \includegraphics[width=0.45\paperwidth]{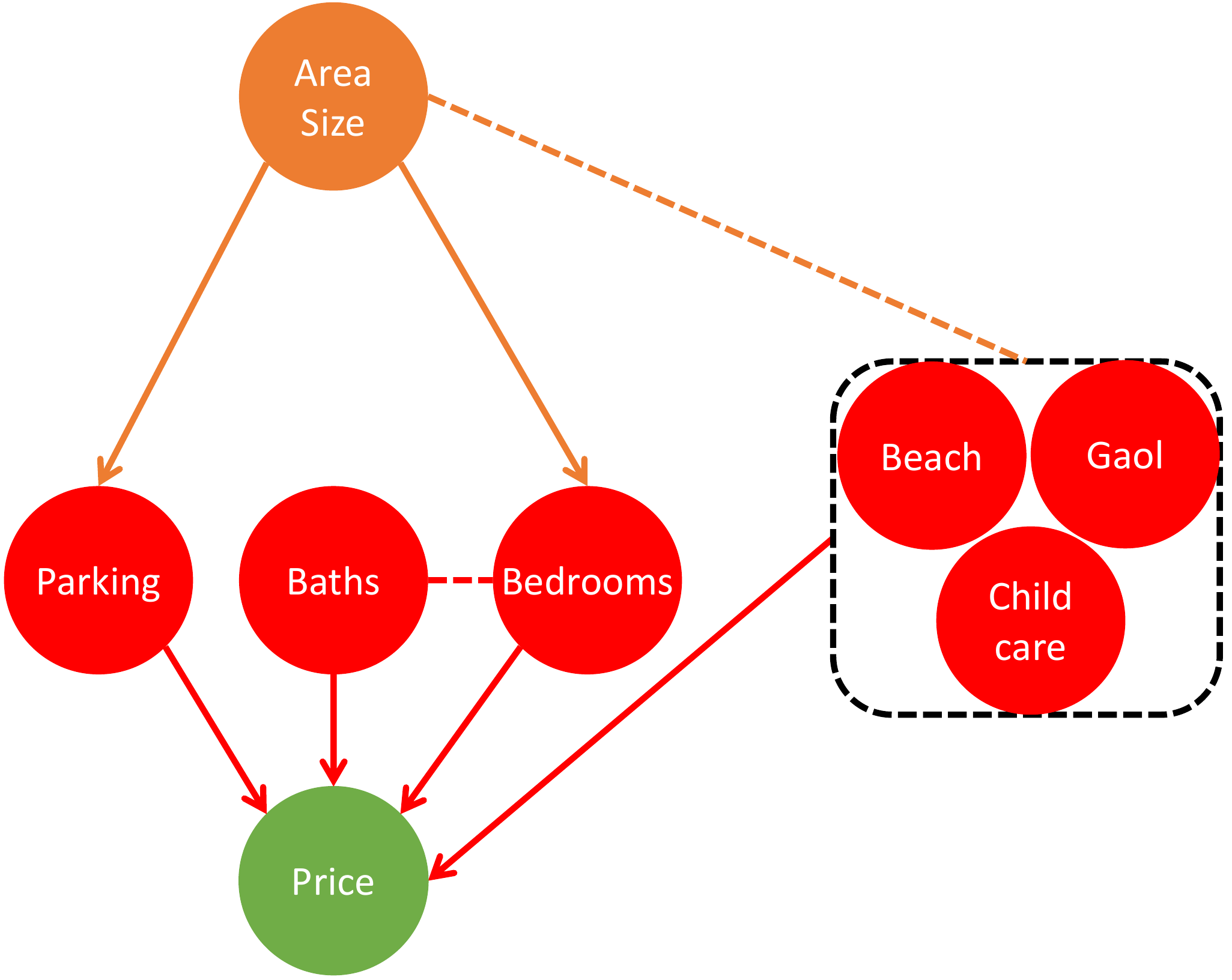}
  \caption{Illustration on the relation of AreaSize to the parents of price.}
  \label{fig:area_size}
\end{figure}

It is worth noting that AreaSize is not selected into the MB of price, which seems counterintuitive. However, this can be explained using Figure~\ref{fig:area_size}. Firstly, area size of houses at different locations are not comparable. A downtown terrace house with small area size can be much more expensive than a house on a large plot 10km from downtown. Hence, ceteris paribus, AreaSize has much more explanatory power on price if we compare houses within a local area or spatial neighbourhood, which requires a spatial statistics technique\footnote{For example, lasso variable selection within a spatial Gaussian kernel.} as opposed to simply controlling for distance to various locations. Secondly, AreaSize measure the area of the land and is originally determined when the land is purchased, which is even earlier than the house construction. This implies that AreaSize is one of the constraints for the construction of parking, baths rooms and bedrooms. Thus, AreaSize is likely a parent of parking and bedrooms and may also be simultaneously determined with location when the land is purchased. Thus, AreaSize does not belong to the MB of price. It is also interesting to note that, unlike parking and bedrooms, the number of baths is not a child of AreaSize and that there is a undirected edge between baths and bedrooms. This finding is data-driven because $\mathrm{corr} \left( \mathrm{Bath}, \mathrm{AreaSize} \right) \approx 0$ while $\mathrm{corr} \left( \mathrm{Bath}, \mathrm{bedrooms} \right)$ is significantly nonzero in the data. This phenomenon is not unexpected. During house design and construction, the number of bathrooms is typically a function of the number of bedrooms (essentially, of expected household size). Given household size and number of bedrooms, there seems to be little incentive to build more bathrooms with more AreaSize. At the end of the day, due to a lack of data on the first-hand house market and new construction, we are unable to infer a graph incorporating construction and land purchase and will not pursue the topic further.

\section{Application of graph estimation: endogeneity detection and instrument selection \label{section:application}}

With the estimated graph in hand, we can begin the instrument selection procedure using Definition~\ref{def:instrument_variable}. The first step to selecting a valid instrument is to ensure there is endogeneity in the graph, otherwise we only waste degrees of freedom.

\subsection{Endogeneity detection using graphs}

Price is endogenous statistically and empirically. Figure~\ref{fig:BN_post} depicts a graph that reflects a statistically dynamic system. The input of the system is the gaol group, house features and ICSEA, rent and mortgage are two outputs, and price is internally determined by the statistical system. As a result, price is highly likely to be endogenous. The endogeneity of price is also supported statistically by variable selection results. For example, by estimating linear and log solar regressions of rent on all the other variables, we have the following estimated regression models,
\begin{align}
  \mathrm{logRent} = \alpha_0 & + \alpha_1 \cdot \mathrm{TotPop} + \alpha_2 \cdot \mathrm{Household\_size} + \alpha_3 \cdot \mathrm{Beach} + \alpha_4 \cdot \mathrm{ChildCare} \notag  \\
  & + \alpha_5 \cdot \mathrm{Gaol} + \alpha_6 \cdot \mathrm{PrimaryHigh}  + \alpha_7 \cdot \mathrm{ICSEA} + \alpha_8 \cdot \mathrm{logPersonInc} \notag   \\
  & + \alpha_9 \cdot \mathrm{logFamInc} + \alpha_{10} \cdot \mathrm{logPrice} + u, \label{OLS:solar_logRent} \\
  \mathrm{Rent} = \gamma_0 & + \gamma_1 \cdot \mathrm {Household\_size} + \gamma_2 \cdot \mathrm{Beach} + \gamma_3 \cdot \mathrm{ChildCare} + \gamma_4 \cdot \mathrm{Gaol} \notag \\
  & + \gamma_5 \cdot \mathrm{PrimaryHigh} + \gamma_6 \cdot \mathrm{Mortgage} + \gamma_7 \cdot \mathrm{ICSEA} + \gamma_8 \cdot \mathrm{FamInc} \notag \\
  & + \gamma_9 \cdot \mathrm{PersonInc} + \gamma_{10} \cdot \mathrm{Price} + u. \label{OLS:solar_Rent}
\end{align}
Since the $p/n$ ratio is almost $1/200$, the solar variable selection results are statistically robust and accurate. Solar includes log(price) and price in (\ref{OLS:solar_logRent}) and (\ref{OLS:solar_Rent}), respectively, implying that price is an very important covariate in the rent regressions and that it is highly likely that rent and price are simultaneously determined. Thus, along with the solar variable selection results in (\ref{OLS:solar}) and (\ref{OLS:solar_log}), we can establish the simultaneous equations models in log terms
\begin{equation}
  \begin{cases}
    \mathrm{logRent} = \alpha_0 & + \alpha_1 \cdot \mathrm{TotPop} + \alpha_2 \cdot \mathrm{Household\_size} + \alpha_3 \cdot \mathrm{Beach} + \alpha_4 \cdot \mathrm{ChildCare} \\
    & + \alpha_5 \cdot \mathrm{Gaol} + \alpha_6 \cdot \mathrm{PrimaryHigh}  + \alpha_7 \cdot \mathrm{ICSEA} + \alpha_8 \cdot \mathrm{logPersonInc} \\
    & + \alpha_9 \cdot \mathrm{logFamInc} + \alpha_{10} \cdot \mathrm{logPrice} + u_1, \\
    \mathrm{logPrice} = \beta_0 & + \beta_1 \cdot \mathrm{logMortgage} + \beta_2 \cdot \mathrm{logRent} + \beta_3 \cdot \mathrm{logFamInc} + \beta_4 \cdot \mathrm{Bedrooms} \\
    & + \beta_5 \cdot \mathrm{Baths} + \beta_6 \cdot \mathrm{Parking} + \beta_7 \cdot \mathrm{Beach} + \beta_8 \cdot \mathrm{Gaol} + \beta_9 \cdot \mathrm{ICSEA} + u_2;
  \end{cases}
\label{eqn:SEM_linear}
\end{equation}
or in linear terms
\begin{equation}
  \begin{cases}
    \mathrm{Rent} = \gamma_0 & + \gamma_1 \cdot \mathrm {Household\_size} + \gamma_2 \cdot \mathrm{Beach} + \gamma_3 \cdot \mathrm{ChildCare} + \gamma_4 \cdot \mathrm{Gaol} \\
    & + \gamma_5 \cdot \mathrm{PrimaryHigh} + \gamma_6 \cdot \mathrm{Mortgage} + \gamma_7 \cdot \mathrm{ICSEA} + \gamma_8 \cdot \mathrm{FamInc} \\
    & + \gamma_9 \cdot \mathrm{PersonInc} + \gamma_{10} \cdot \mathrm{Price} + u_1, \\
    \mathrm{Price} = \delta_0 & + \delta_1 \cdot \mathrm{Mortgage} + \delta_2 \cdot \mathrm{Rent} +
    \delta_3 \cdot \mathrm{FamInc} + \delta_4 \cdot \mathrm{Bedrooms} \\
    & + \delta_5 \cdot \mathrm{Baths} + \delta_6 \cdot \mathrm{Parking} + \delta_7 \cdot \mathrm{Beach} + \delta_8 \cdot \mathrm{Gaol} + \delta_9 \cdot \mathrm{ICSEA} + u.
  \end{cases}
  \label{eqn:SEM_log}
\end{equation}

The simultaneous determination of rent, price and mortgage is also empirically intuitive. Before bidding on a house, the buyer needs to estimate the upper bound of mortgage that a bank will offer and, if the purchase is for investment purposes, how much rent the property will return. Before a bank decides on a mortgage application, it typically first gets a valuation of the house price and the expected rent. Similarly, rent is typically related to the price of house and monthly mortgage amount.\footnote{In a competitive market with zero transactions costs, of course, rent would exactly cover the mortgage, which itself would be equal to the price of the house.}

\subsection{Instrument selection using graphs}

\begin{figure}[H]
  \centering
  \includegraphics[width=0.5\paperwidth]{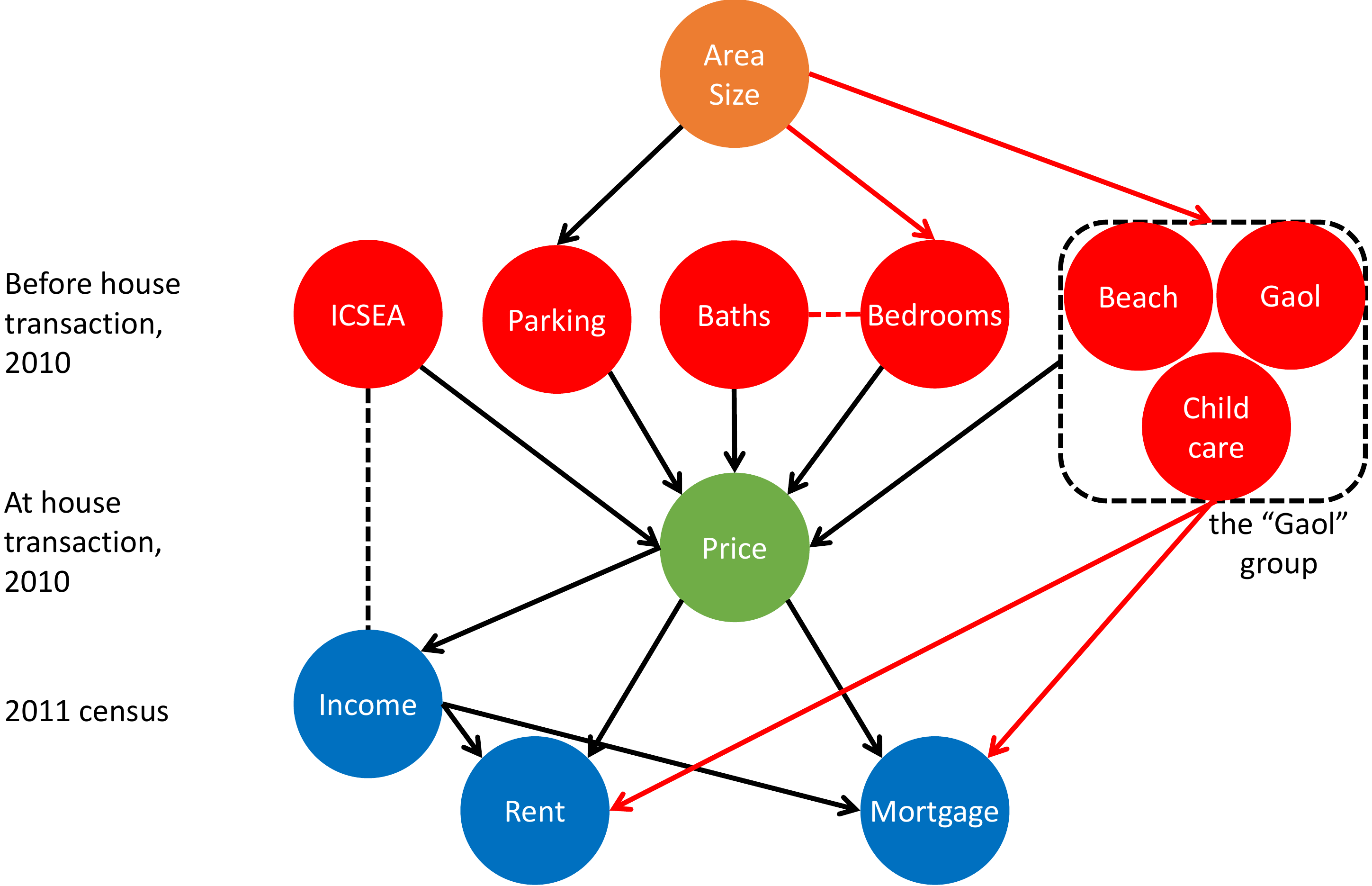}
  \caption{Backdoor effect from Bath to Rent (highlighted in red).}
  \label{fig:BE_house}
\end{figure}

Given price is endogenous, we need to find a valid instrument in the rent regression. We focus on the regression analysis of rent in this paper, but the mortgage analysis proceeds along the same lines. A valid instrument must satisfy Definition~\ref{def:instrument_variable} and non-existence of a BE (Figures~\ref{fig:instrument} and \ref{fig:not_instrument1}). Using Figure~\ref{fig:BN_post}, we directly uncover 3 instrumental variables: baths, bedrooms and parking, all of which satisfy Definition~\ref{def:instrument_variable} only if we control the Gaol group variables. As shown in Figure~\ref{fig:BE_house}, if we fail to control the Gaol group variables, a backdoor effect will be constructed between Rent and any of\{baths, bedrooms, parking\} as follows :due to the existence of the confounder AreaSize, Baths, for example, will be unconditionally correlated to Bedrooms, which is further unconditionally correlated to the Gaol group variables due to the confounder AreaSize. Since the Gaol group variables --- variables to represent the house location --- causes the change of the rent, a BE is constructed. This violates Definition~\ref{def:instrument_variable} by allowing the change of Baths to affect the change Rent even though you control the house price. Under such circumstance, Baths cannot be an instrument for the causal effect from Price to Rent since it represent two effects: the causal effect from Price to Rent and the backdoor effect mentioned above.Hence, the Gaol group variables must be controlled, meaning that they have to be included into the instrument variable regression of rent.

With Gaol group variables controlled, Figure~\ref{fig:Backdoor_bath} and Table~(\ref{Table:Backdoor}) confirm that all 3 variables can only affect rent through price, which satisfies G2 in Definition~\ref{def:instrument_variable}. Yet these two conditions are for a general statistical dynamic system. The score-based learning method we use only requires the subgaussian distribution of each variable, which permits causation among variables to be any nonlinear form.\footnote{For nonlinear graph estimation with endogeneity bias correction, we need dependence measures such as mutual information and the Hilbert-Schmidt independence criterion, both of which would require enormous computation loads in our high-dimensional database. Hence, we skip this topic in this paper.} However, classic endogeneity analysis in regression requires linearity among all the variables. To ensure these 3 variables fit a system of linear regression equations, we need to check the correlation between rent and the instrument for price, shown in Table~\ref{table:iv_correlation}.

\begin{table}[H]
  \caption{Correlation table for the instruments, endogenous variable, and rent}
  \label{table:iv_correlation}
  \begin{tabular}{l.....}
    \toprule
             & \multicolumn{1}{c}{Baths}
             & \multicolumn{1}{c}{Parking}
             & \multicolumn{1}{c}{Bedroom}
             & \multicolumn{1}{c}{Price}
             & \multicolumn{1}{c}{logPrice} \\
    \midrule
        Rent     & 0.19  & 0.069   & 0.061   & 0.34  &          \\
        logRent  & 0.16  & 0.043   & 0.023   &       & 0.32     \\
        Price    & 0.52  & 0.34    & 0.46    &       &          \\
        logPrice & 0.57  & 0.41    & 0.60    &       &          \\
    \bottomrule
  \end{tabular}
\end{table}

Table~\ref{table:iv_correlation} clearly shows that bedrooms and parking do not fit the linear system as instruments for price: $\mathrm{corr} \left( \mathrm{logRent}, \mathrm{Parking} \right)$ and $\mathrm{corr} \left( \mathrm{logRent}, \mathrm{bedrooms} \right)$ are too weak. As a result, even though the correlations between these variables and logPrice are quite high, the predicted value of logPrice using these two variables cannot explain enough of the variation in logRent using 2SLS or other IV regressions, which suggests the possibility of weak instruments and in turn may lead to the wrong signs, values of the regression coefficients and misleading interpretation. These concerns are confirmed in Tables~\ref{table:IV_2LSL_log} and \ref{table:IV_2LSL_linear}.

\begin{table}[H]
  \caption{logRent regressions: 2SLS coefficients and t-values}
  \label{table:IV_2LSL_log}
  \begin{tabular}{r....}
    \toprule
     &     \multicolumn{1}{c}{OLS} & \multicolumn{3}{c}{2SLS} \\ \cline{3-5}
     & & \multicolumn{1}{c}{Baths} & \multicolumn{1}{c}{Bedrooms} & \multicolumn{1}{c}{Parking} \\
    \midrule
    const           & 1.87     & 1.60       & 2.2169        & 1.91         \\
                    & (17.34)  & (11.10)    & (17.92)       & (11.77)      \\
    TotPop          & 0.0002   & 0.0002     & 0.0002        & 0.0002       \\
                    & (11.31)  & (11.63)    & (10.40)       & (10.84)      \\
    household\_size & 0.31     & 0.29       & 0.32          & 0.31         \\
                    & (27.01)  & (23.60)    & (27.22)       & (23.31)      \\
    Beach           & -2.35    & -2.21      & -2.55         & -2.38        \\
                    & (-14.04) & (-12.65)   & (-14.67)      & (-12.80)     \\
    ChildCare       & -1.67    & -1.57      & -1.81         & -1.69        \\
                    & (-12.31) & (-11.11)   & (-12.90)      & (-11.42)     \\
    Gaol            & 1.23     & 1.19       & 1.28          & 1.24         \\
                    & (7.09)   & (6.82)     & (7.35)        & (7.02)       \\
    PrimaryHigh     & -1.30    & -1.37      & -1.22         & -1.29        \\
                    & (-7.81)  & (-8.13)    & (-7.21)       & (-7.63)      \\
    ICSEA           & 0.0004   & 0.0004     & 0.0005        & 0.0004       \\
                    & (5.81)   & (5.00)     & (6.76)        & (5.77)       \\
    logPersonInc    & 0.50     & 0.51       & 0.49          & 0.50         \\
                    & (17.46)  & (17.66)    & (17.10)       & (17.31)      \\
    logFamInc       & -0.04    & -0.07      & -0.0090       & -0.0409      \\
                    & (-1.84)  & (-2.76)    & (-0.35)       & (-1.49)      \\
    logPrice        & 0.0053   & 0.0410     & -0.0428       & -0.0014      \\
                    & (0.72)   & (2.80)     & (-3.90)       & (-0.07)      \\
    \midrule
    $p$         & & & & \\
    $R^2$       & 0.3650    & 0.3632     & 0.3618        & 0.3649       \\
    $\bar{R}^2$ & 0.3645    & 0.3627     & 0.3612        & 0.3644       \\
    $n$         & 11,796    & 11,796     & 11,796        & 11,796        \\
    \bottomrule
  \end{tabular}
\end{table}

Table~\ref{table:IV_2LSL_log} shows the 2SLS estimates based on the log regressions. The covariates in the table are chosen by data-driven variable selection algorithms. In Table~\ref{table:IV_2LSL_log}, we use baths, bedrooms, and parking as logPrice instruments, respectively. Due to the endogeneity, OLS clearly underestimates the marginal effect of logPrice where it is not significant. In the baths 2SLS, the logPrice coefficient is 7 times larger and the t-value of logPrice is 3 times larger than for the corresponding OLS coeffcient, so that it is now significant. These results clearly show the bias correction effect of 2SLS using baths. By contrast, due to the weak correlation between parking, bedrooms and logRent, both the parking and bedrooms 2SLS move the OLS coefficient of logPrice in the wrong direction, giving the wrong interpretation that higher house prices are associated with lower rent. Also, logPrice in the parking 2SLS is even less significant than in OLS.

\begin{table}[H]
  \caption{Rent regressions: 2SLS coefficients and t-values}
  \label{table:IV_2LSL_linear}
  \begin{tabular}{r....}
    \toprule
     &     \multicolumn{1}{c}{OLS} & \multicolumn{3}{c}{2SLS} \\ \cline{3-5}
     & & \multicolumn{1}{c}{Baths} & \multicolumn{1}{c}{Bedrooms} & \multicolumn{1}{c}{Parking} \\
    \midrule
    const            & -74.336   & -50.537   & -116.40     & -80.687     \\
                     & (-2.2952) & (-1.4809) & (-3.3948)   & (-2.2233)   \\
    household\_size  & 105.60    & 102.56    & 110.95      & 106.40      \\
                     & (25.829)  & (23.629)  & (25.427)    & (23.053)    \\
    Beach            & -841.34   & -795.41   & -922.53     & -853.60     \\
                     & (-12.933) & (-11.478) & (-13.413)   & (-11.477)   \\
    ChildCare        & -574.23   & -549.46   & -618.02     & -580.84     \\
                     & (-10.151) & (-9.3269) & (-10.576)   & (-9.5272)   \\
    Gaol             & 329.80    & 315.15    & 355.71      & 333.72      \\
                     & (4.5646)  & (4.2765)  & (4.8245)    & (4.4669)    \\
    PrimaryHigh      & -331.82   & -343.79   & -310.66     & -328.62     \\
                     & (-4.7021) & (-4.8783) & (-4.3918)   & (-4.6602)   \\
    Mortgage         & 0.0154    & 0.0126    & 0.0203      & 0.0161      \\
                     & (3.4684)  & (2.5347)  & (4.0522)    & (3.0657)    \\
    ICSEA            & 0.0004    & 0.0004    & 0.0005      & 0.0004      \\
                     & (5.8185)  & (5.0046)  & (6.7625)    & (5.7714)    \\
    ICSEA            & 0.0855    & 0.0678    & 0.1169      & 0.0902      \\
                     & (2.9184)  & (2.3028)  & (3.9957)    & (2.9591)    \\
    FamInc           & -0.0012   & -0.0034   & 0.0029      & -0.0006     \\
                     & (-0.1717) & (-0.5222) & (0.4400)    & (-0.0846)   \\
    Inc              & 0.2250    & 0.2276    & 0.2203      & 0.2243      \\
                     & (15.635)  & (15.970)  & (15.410)    & (15.688)    \\
    Price            & 1.95e-05  & 3.04e-05  & 1.98e-07    & 1.66e-05   \\
                     & (3.8921)  & (5.3430)  & (0.0444)    & (2.4782)    \\
    \midrule
    $p$         & & & & \\
    $R^2$       & 0.3462    & 0.3444    & 0.3407      & 0.3461      \\
    $\bar{R}^2$ & 0.3457    & 0.3439    & 0.3401      & 0.3455      \\
    $n$         & 11,974    & 11,974    & 11,974      & 11,974      \\
    \bottomrule
  \end{tabular}
\end{table}

Table~\ref{table:IV_2LSL_linear} shows the 2SLS results in the linear regressions, which are similar to Table~\ref{table:IV_2LSL_log}. OLS still underestimates the marginal effect of price. The price coefficient in the baths 2SLS is around 50\% larger than the corresponding OLS coefficient. The t-value of price in the baths 2SLS is 40\% larger than the OLS t-value. Similar to Table~\ref{table:IV_2LSL_log}, the parking and bedrooms 2SLS coefficient of price and corresponding marginal effect are even smaller than the corresponding marginal effect in the endogenous OLS.

Finally, to double-check the validity of each instrument in 2SLS, we implement 4 traditional instrument tests and report results in Table~\ref{table:instrument_test}.

\begin{table}[H]
  \centering
  \caption{Classical tests of instruments in the linear and log regressions}
  \label{table:instrument_test}
  \begin{tabular}{ll......}
    \toprule
     & & \multicolumn{3}{c}{log} & \multicolumn{3}{c}{linear} \\ \cline{3-8}
    \\ [-8pt]
     & & \multicolumn{1}{r}{Baths}
       & \multicolumn{1}{r}{Bedrooms}
       & \multicolumn{1}{r}{Parking}
       & \multicolumn{1}{r}{Baths}
       & \multicolumn{1}{r}{Bedrooms}
       & \multicolumn{1}{r}{Parking} \\
    \midrule
    \multirow{2}{*}{\parbox{2.5cm}{Durbin test}}
     & $\chi^2_1$ & 13.0337 & 49.5049 & 0.2538 & 8.8532 & 29.8844 & 0.2795 \\
     & p-value & 0.0003 & 0.0000 & 0.6144 & 0.0029 & 0.0000 & 0.5970 \\
    \midrule
    \multirow{2}{*}{\parbox{2.5cm}{Wu-Hausman test}}
     & $F_{1,11784}$ & 13.0348 & 49.6629 & 0.2535 & 8.8508 & 29.9291 & 0.2793 \\
     & p-value & 0.0003 & 0.0000 & 0.6146 & 0.0029 & 0.0000 & 0.5972 \\
    \midrule
    \multirow{2}{*}{\parbox{2.2cm}{Wooldridge regression test}}
     & $\chi^2_1$ & 9.1282 & 31.6738 & 0.1725 & 4.5087 & 14.0788 & 0.1427 \\
     & p-value & 0.0025 & 0.0000 & 0.6779 & 0.0337 & 0.0002 & 0.7056 \\
    \midrule
    \multirow{2}{*}{\parbox{2.5cm}{Wooldridge score test}}
     & $\chi^2_1$ & 9.2345 & 33.7213 & 0.1734 & 4.5083 & 14.8951 & 0.1433 \\
     & p-value & 0.0024 & 0.0000 & 0.6771 & 0.0337 & 0.0001 & 0.7050 \\
    \bottomrule
  \end{tabular}
\end{table}

With p-values less than 1\%, all 4 tests confirm that baths significantly corrects the endogeneity bias on the logPrice marginal effect in 2SLS. Consistent with our intuition, graph estimation and MB selection successfully accomplish endogeneity detection, instrument validation and selection. In the linear models, the p-values of baths increase marginally, yet stay below 5\%. This is as expected because we do not log-transform dollar-measured variables, which leaves them with long, heavy tails and highly likely non-subgaussian. Consistent with previous concerns, the p-values for all tests in the parking 2SLS are well above 5\% (linear and log) due to the weak correlation between the instrument and the response variable. This shows that graph estimation and MB selection work wells on instrument validity in our data. For similar reasons, bedrooms also alters the price marginal effect in the wrong direction, making price insignificant and logPrice the wrong sign interpreted. In this case, the low p-values for the bedrooms 2SLS is due to the size of the misscorrection and do not imply validity of the instrument.

\subsection{Sanity check of the graph learning and interpretation on the validity of instruments}

We also need check the reliability of the learning result and investigate its empirical appropriateness. Unlike the house pricing regression, the rent regression only returns an $R^2$ of around 40\%. The low $R^2$ suggests that some variation of rent is not specified as a linear model. As a result, we need to carefully check the model.

In the determination of house prices, it is intuitive that bedrooms directly causes a change in price: more bedrooms require extra construction cost. But the finding that bedrooms does not cause rent directly may seem counterintuitive. Based on 2011 census and the house leasing data, the majority of leasing demand comes from university and international students, young professionals and couples without children. Due to the short house supply and great leasing demand, a great number of landlords lease their rooms to different tenants or couples, each of which only occupies one room. Hence, the number of bedrooms is not relevent to the tenants as long as there is one avaiable. Even worse, a great number of landlords lease their rooms via room-sharing: a house can accommodate more tenants via partitioning bedrooms or turning common living spaces (e.g., lounges, dining rooms, etc.) into bedrooms. While illegal (though hard to check), partitioning rooms can greatly increase the number of tenants in a house and, hence, the rental income, which makes room-sharing quite popular in Sydney leasing market.\footnote{See, for example, the house releasing report of Rent.com.au at \url{https://www.rent.com.au/blog/room-sharing-overcrowding}.} As a result, bedrooms does not accurately reflect leasing capacity and, hence, does not cause rent directly, explaining why the correlation between bedrooms and logRent is low. Furthermore, the negative coefficient of $\log \left( \mathrm{Price} \right)$ in the Bedroom IV equation may be the classical Simpson's paradox. Since we cannot observe the number of shared rooms in a house, we cannot control this variable when doing causal inference. Controlling other variables and the number of shared rooms, we may still find a positive correlation between Bedroom and Price. For details of modelling the latent `number of shared rooms', see Appendix.

The number of parking directly causes the construction cost of the house and, further, the price. A reason similar to Bedroom can be found for why parking does not cause rent directly. The city council of Sydney issues permits to all local residents for street parking without any time limit, meaning that the parking variable does not accurately measure the parking spaces a house tenant can access. Similar to bedrooms, baths also causes price directly and rent indirectly, which seem quite natural. However, in the second-hand house market in Sydney, baths may reflect whether a house is recently refurbished or constructed. In our 2010 database, we find that more than 60\% of the transacted houses (many of which are terrace houses or town houses close to the CBD) are more than 80 years old. The houses designed at that time typically have only one bathroom, regardless of the number of bedrooms. Thus, a house with two or more bathrooms is likely either a newly constructed house or a recently refurbished house. Hence, baths is possibly a strong indicator of house quality and explains why the correlation between baths and logRent is much larger than $\mathrm{corr} \left( \mathrm{baths}, \mathrm{logRent} \right)$. Besides, unlike Bedroom, Baths is unlikely to be affected by the room-sharing problem, since it is much more difficult and troublesome to rebuild a living room into a bath room. This further explains why baths is a valid instrument for logPrice in 2SLS.

\section{Conclusion}

In this paper, we demonstrate the performance of solar variable selection with empirical data that have severe multicollinearity and, hence, severe grouping effects. As a competitor to solar, lasso is more sensitive to the grouping effect and returns unreliable variable-selection results. While more robust to the grouping effect than lasso, CV-en loses all sparsity in variable selection. By contrast, solar returns a stable and sparse variable selection and illustrates superior robustness to the grouping effect.

To be added:
\begin{itemize}
  \item summary of graph learning application.
  \item caveats.
  \item topics for further research (including potential data applications).
\end{itemize}

%%%%%%%%%%%%%%%%%%%%%%%%%%%%%%%%%%
%%%%%%%%%%% References %%%%%%%%%%%
%%%%%%%%%%%%%%%%%%%%%%%%%%%%%%%%%%

\newpage

\bibliographystyle{elsarticle-harv}
\bibliography{solar_refs,bayes_net_refs}

\end{document}